# G2-MonoDepth: A General Framework of Generalized Depth Inference from Monocular RGB+X Data

Haotian Wang, Meng Yang*, *Member, IEEE*, and Nanning Zheng*, *Fellow, IEEE*

**Abstract**—Monocular depth inference is a fundamental problem for scene perception of robots. Specific robots may be equipped with a camera plus an optional depth sensor of any type and located in various scenes of different scales, whereas recent advances derived multiple individual sub-tasks. It leads to additional burdens to fine-tune models for specific robots and thereby high-cost customization in large-scale industrialization. This paper investigates a unified task of monocular depth inference, which infers high-quality depth maps from all kinds of input raw data from various robots in unseen scenes. A basic benchmark G2-MonoDepth is developed for this task, which comprises four components: (a) a unified data representation RGB+X to accommodate RGB plus raw depth with diverse scene scale/semantics, depth sparsity ([0%, 100%]) and errors (holes/noises/blurs), (b) a novel unified loss to adapt to diverse depth sparsity/errors of input raw data and diverse scales of output scenes, (c) an improved network to well propagate diverse sparse scales from input to output, and (d) a data augmentation pipeline to simulate all types of real artifacts in raw depth maps for training. G2-MonoDepth is applied in three sub-tasks including depth estimation, depth completion with different sparsity, and depth enhancement in unseen scenes, and it always outperforms SOTA baselines on both real-world data and synthetic data.

——————————— ◆ ———————————

## 1 INTRODUCTION

Monocular depth inference has long been a fundamental problem for robots to perceive real (unseen) scenes, such as its downstream applications of 3D object detection [54], 3D odometry [53], 3D reconstruction [32], human pose estimation [63], and robot localization [15]. In different applications, various robots may be equipped with a single camera or a camera plus a depth sensor of any type such as Time-of-Flight (ToF), Structured-Light (SL), 4/16/64/128-line LiDAR, and may be located in various scenes with different scales such as indoor bedroom, kitchen, outdoor traffic, urban, or landscape scenes. Recent advances derived various sub-tasks of monocular depth inference depending on the specific configuration of sensors aforementioned and scenes with different scales in the field of machine vision [2] [3] [5] [6] [7] [8]. However, task-specific models induce additional burdens of computing and time to retrain or fine-tune models for specific robots, and thereby lead to high costs of customized hardware or chips in large-scale industrialization.

To bridge this gap, we investigate a *unified task* of monocular depth inference for various robots in various scenes. In our task, high-quality depth maps are inferred from all kinds of input raw data captured by different configurations of sensors aforementioned in unseen output scenes with different scales. We summarize two main challenges to develop a model for our unified task.

### 1.1 Generality on Input Data Types

Robots equipped with a single camera and an optional depth sensor of any type may capture three types of raw data with varying valid depth values, namely, *a single RGB image*, *an RGB image plus a sparse depth map*, and *an RGB image plus a dense depth map*. Monocular depth inference from these types of raw data has derived three individual sub-tasks in the literature, including *depth estimation* [3], *depth completion with different sparsity* [4], and *depth enhancement* [5]. Depth estimation infers a high-quality depth map from an RGB image captured by a single camera [10]. Depth completion infers a high-quality depth map from an RGB image plus a sparse depth map with holes captured by 4/16/64/128-line LiDAR [11] or SL [12]. Depth enhancement recovers a high-quality depth map from an RGB image plus a dense yet distorted depth map with blurs (e.g., low-resolution) and noises captured by SL, ToF [30] or estimated by stereo matching [9]. In the past decades, plenty of models have been developed for these three individual sub-tasks [1] [3] [31] [39] [40] [51], which are specifically designed for one type of raw data aforementioned and are incompatible with the other two types. Furthermore, existing models of the same data type cannot be well applied to raw data with different sparsity

• *Haotian Wang, Meng Yang, and Nanning Zheng are with the Institute of Artificial Intelligence and Robotics, Xi'an Jiaotong University, Xi'an 710049, China. E-mails: wht_sxchina@stu.xjtu.edu.cn; {mengyang, nnzheng}@mail.xjtu.edu.cn.*
• *Code is available at https://github.com/Wang-xjtu/G2-MonoDepth*

***\*\*\*Please provide a complete mailing address for each author, as this is the address the 10 complimentary reprints of your paper will be sent***

*Please note that all acknowledgments should be placed at the end of the paper, before the bibliography (note that corresponding authorship is not noted in affiliation box, but in acknowledgment section).*

(e.g., 4/16/64/128-line) and errors (e.g., holes, blurs, and noises). Recently, a few efforts were made to accommodate more raw data in a single model. For example, researchers introduced additional modules into models of individual sub-tasks to enable two types of raw data [50] [55]. Some models are trained on simulated data to adapt to raw data with different sparsity and errors [8] [57] [58]. However, no prior work well investigates the inherent correlation of these input raw data as well as the three "individual" sub-tasks in the literature. **In this paper, we analyze that these individual sub-tasks of monocular depth inference from all these raw data are essentially the same. Thereby, we explore a general framework of monocular depth inference for our unified task to naturally accommodate all the three types of input raw data as well as diverse depth sparsity and errors.**

### 1.2 Generalization on Output Scene Scales

Robots may be located in unseen scenes with different *scene scales* as well as *scene semantics*. *Traditional depth inference* usually trains and tests models in a single dataset with similar scene scales and semantics such as depth estimation [44], depth completion [21], and depth enhancement [13]. However, these models cannot be well generalized to unseen scenes in practical applications. Consequently, two sub-tasks are derived in recent years for generalized depth inference including *scale-invariant depth inference* [7] and *zero-shot depth inference* [16]. First, scene scales may vary drastically in different scenes or even in similar scenes [7]. For example, the scales of indoor bedroom scenes often range within 10 meters, whereas the scales of outdoor traffic scenes are often up to 80 meters. *Scale-invariant depth inference* generally improves the generalization ability by normalizing unknown scales in the sub-task of depth estimation [6] [17] [19]. Second, the semantics of various scenes may be completely different, such as indoor bedroom with manmade furniture (e.g., NYUv2 [14]) and outdoor traffic with cars and pedestrians (e.g., KITTI [15]). *Zero-shot depth inference* usually improves the generalization ability for unknown semantics by training on diverse data [6] [16] [19]. Recently, some methods such as MiDas [16] attempted to jointly consider scene scales/semantics in the sub-task of depth estimation [17] [19]. **To our knowledge, the generalization issue of depth inference was partly studied in the sub-task of depth estimation to generate relative depth maps. However, it remains a challenge in sub-tasks of depth completion [21] [39] and depth enhancement [13] [41], which require to generate absolute depth maps. In this paper, we further explore a generalized framework of monocular depth inference for our unified task to fully accommodate diverse scales and semantics of output scenes.**

### 1.3 Contributions

In this paper, we investigate a unified task of monocular depth inference, which aims to infer a high-quality depth map from input raw data with diverse depth sparsi-ty/errors in unseen output scenes with diverse scales/semantics. A basic benchmark **G2-MonoDepth** is developed for this unified task, which mainly comprises four components, including a unified data representation RGB+X, a unified G2-MonoDepth loss, an improved network ReZero U-Net, and a data augmentation pipeline.

**RGB+X data representation** is used to uniformly represent all types of input raw data in our solution. It refers to RGB images with diverse scene scales and semantics *plus* optional raw depth maps with diverse sparsity (i.e., arbitrary [0%, 100%][1] valid pixels) and errors (i.e., holes, noises, and blurs). Our RGB+X data representation fully accommodates all existing sub-tasks of monocular depth inference in unseen scenes. For example, depth estimation, depth completion with different sparsity, and depth enhancement require the inputs of an RGB image plus 0%, (0%, 100%), and 100% valid depth pixels, separately.

**G2-MonoDepth loss** establishes a constraint between inferred depth map and ground-truth (GT) depth map conditioned upon monocular RGB+X data. In our solution, the loss is required to adapt to not only diverse depth sparsity/errors of input raw data but also diverse scales of output scenes. Our G2-MonoDepth loss is mathematically derived from a relative relation from RGB image and an absolute relation from X map. On the one hand, a relative relation is constrained between inferred depth map and GT depth map conditioned upon a single RGB image. The reason is that the scale of inferred depth map is unknown because of the scale-ambiguity problem of a single RGB image. On the other hand, an absolute relation is constrained for valid depth pixels between inferred depth map and GT depth map conditioned upon an X map. The reason is that X map only provides the intensities of valid depth pixels, whereas the intensities of other depth pixels are unknown. We analyze that our G2-MonoDepth loss naturally accommodates three types of widely-used loss functions in monocular depth inference, including L1 or L2 [4] [5] [10], scale-invariant loss [7] [17] [50], and affine-invariant loss [16] [45].

**ReZero U-Net** improves the generalization ability of our unified model in unseen scenes with diverse scales. In our G2-MonoDepth loss, inferred depth maps are constrained by not only the relative relation from RGB image but also absolute scene scales from X map. However, we find that the widely-used Batch Normalization (BN) [18] in various network architectures may hinder the scale propagation from the input of RGB+X data to the output of inferred depth maps on training datasets with diverse scene scales. It is indeed one of the reasons that existing models of depth inference cannot be well generalized to unseen scenes of different scales. In our solution, we simply address this problem by fully removing BN and adding ReZero [25] in the basic blocks of our network architecture. It ensures that our network well propagates absolute scene scales from input to output. Notably, we adopt a basic U-Net [60] rather than other complicated architectures [7] [44] [45] [49] for low complexity and easy

---

[1] The range [0%, 100%] contains all values from 0% to 100%, which denotes the percentage of valid pixels in raw depth maps. The range (0%, 100%) contains all values in the range [0%, 100%] except for 0% and 100%.

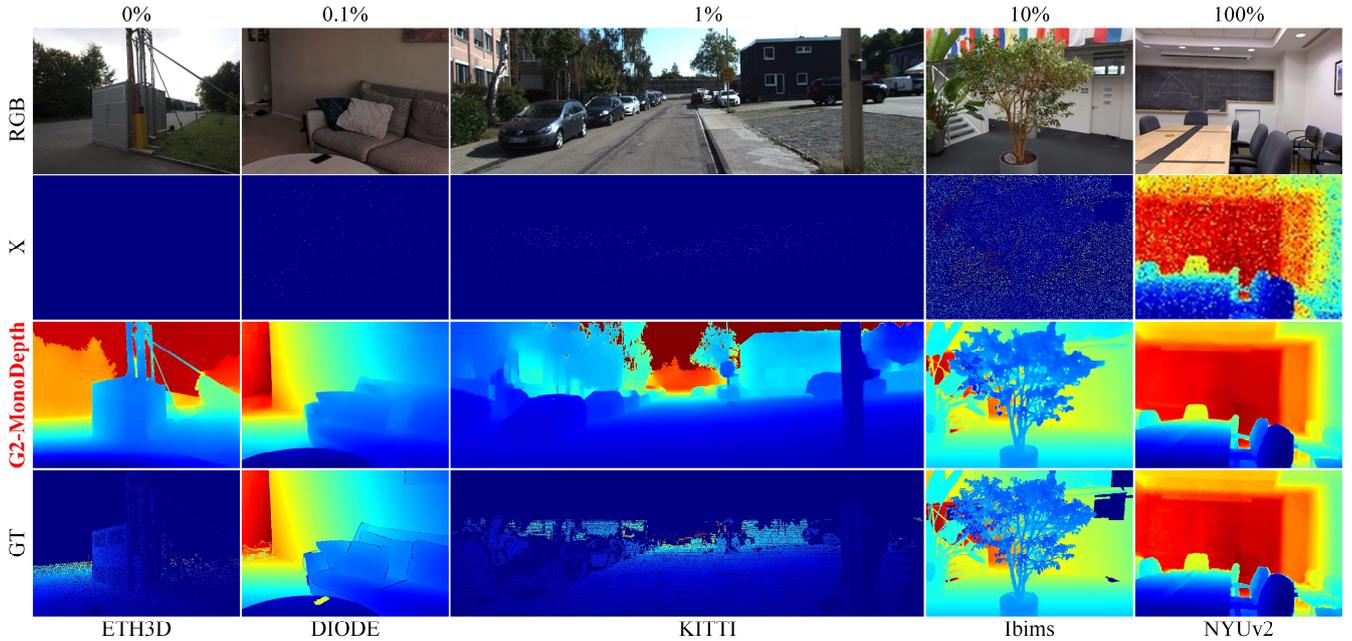

Fig. 1. Examples of inferred depth maps from monocular RGB+X data in unseen scenes with different scene scales/semantics and depth sparsity/errors.

implementation of the benchmark.

**Data augmentation pipeline** well simulates all types of real artifacts in raw depth maps from all kinds of depth sensors in practical scenarios. Though a large quantity of public RGB-D datasets has been released in recent years, most of them do not provide raw depth maps with GT labels [65]. Though a few datasets provide raw depth maps with GT labels such as NYUv2 and KITTI, sparsity/errors of raw depth maps are not diverse enough. It is hard to collect a large-scale RGB-D dataset with diverse scene scales/semantics and depth sparsity/errors, which includes RGB images, GT depth maps, and raw depth maps simultaneously. Our data augmentation pipeline provides an optional way for this problem. We generate an augmented training dataset from Matterport3D [32], HRWSI [19], vKITTI [33], and a new dataset UnrealCV based on the pipeline. It could well improve the generalization ability of our unified model in real scenes.

**Experiments.** Our unified model is applied in the three sub-tasks of depth estimation [16] [17] [19] [38], depth completion with different sparsity [11] [12] [21] [39], and depth enhancement [5] [13] [40] [41] in unseen scenes. Our model is trained on the our augmented training dataset and comprehensively tested on seven unseen real-world datasets including Ibims [34], KITTI [15], NYUv2 [14], DIODE [35], ETH3D [36], Sintel [37], and RealSense [2]. A total of twelve state-of-the-art (SOTA) models of monocular depth inference are used as the baselines for the three sub-tasks. The results demonstrate that our model always outperforms the SOTA baselines of all these sub-tasks on both real-world datasets and augmented datasets, though a basic backbone with low complexity is adopted in our benchmark. Fig. 1 shows several examples of inferred depth maps from monocular RGB+X data in unseen datasets with different scene scales/semantics and depth sparsity/errors.

The main contributions are summarized as follows.

(a) We investigate a unified task of monocular depth inference that infers high-quality depth maps from input raw data with varying depth sparsity/errors in unseen output scenes with varying scales/semantics. We analyze that the sub-tasks of depth estimation, depth completion with different sparsity, and depth enhancement from different types of raw data are essentially the same.

(b) We develop a basic benchmark G2-MonoDepth for this unified task including four components: (a) a unified data representation RGB+X to accommodate all kinds of input raw data with diverse scene scale/semantics and depth sparsity/errors, (b) a novel unified loss to adapt to diverse depth sparsity/errors of input raw data and diverse scales of output scenes, (c) an improved ReZero U-Net to well propagate diverse scene scales from input to output, and (d) a data augmentation pipeline to simulate all types of real artifacts in raw depth maps for training.

(c) G2-MonoDepth always achieves the best quantitative and visual results in all the three sub-tasks of depth estimation, depth completion with different sparsity, and depth enhancement in both unseen real-world data and synthetic data compared with SOTA baselines.

## 2 RELATED WORK

### 2.1 Depth Inference with Different Input Data

Monocular depth inference mainly handles three types of input raw data including a single RGB image, an RGB image plus a sparse depth map, and an RGB image plus a dense depth map. These types of input data derived three sub-tasks including depth estimation, depth completion, and depth enhancement in the literature. **Depth estimation** infers an absolute depth map from a single RGB image in familiar scenes at early time [1] [3] [10] [44]. To address scale-ambiguity problem of single RGB image [7], some methods inferred a relative depth map with normalized scales in unseen scenes [6] [7] [16] [17] [19] [38] [59]. **Depth completion** generally infers an absolute depth map from an RGB image and a sparse depth map with holes. Recent models achieved great success in similar scenes [4] [11] [31] [39] [48], such as NYUv2 [14] and KIT-TI [15]. However, depth completion in unseen scenes remains a challenge in recent years [2] [12]. **Depth enhancement** infers a high-quality depth map from an RGB image and a dense depth map with blurs (e.g., low-resolution) and noises. Numerous models have been proposed in recent years including depth super-resolution [5] [40] [41], depth denoising [51] [52], and depth recovery [13] [20] [64] [65], among which depth super-resolution is the most popular one. However, the generalization issue of learning-based models is not well studied in this subtask compared to traditional filter-based methods [29] [43]. Recently, a few models jointly handled the sub-tasks of depth estimation and depth completion by embedding additional modules into their networks, such as the auxiliary module [50] and the MAP module [55]. Because most RGB-D datasets do not provide raw depth maps with GT labels, or the sparsity and errors of raw depth maps are not diverse enough, many models are trained on simulated data with different types of manual artifacts such as sparsity [4] [12] [21], noises [5] [13] [40], blurs [51] [56], and holes [57] [58], or learned artifacts with Generative Adversarial Network (GANs) [8] [42]. However, no prior work well investigated the inherent correlation of these input raw data as well as the three "individual" sub-tasks in the literature.

## 2.2 Depth Inference in Diverse Output Scenes

Monocular depth inference in unseen output scenes needs to adapt to diverse scene scales and semantics. The generalization issue of monocular depth inference is partly investigated by the sub-tasks of zero-shot depth inference and scale-invariant depth inference in recent years. **Scale-invariant depth inference** infers relative depth maps to ignore the effect of diverse scene scales. Due to scale-ambiguity problem [7], the scales of output scenes are completely unknown given a single RGB image [19] or partly unknown given an RGB image plus a sparse depth map. Unknown scales of output scenes often are normalized by loss functions, such as ranking loss [6], scale-invariant error [7], multi-scale scale-invariant gradient loss [17], and affine-invariant loss [16]. Notably, this paper established a unified loss to naturally accommodate three types of widely-used loss functions, which will be analyzed in Section 3. **Zero-shot depth inference** improves the generalization ability of depth inference to diverse semantics in unseen scenes. It is often realized by collecting training datasets with diverse semantics, such as DIW [6], MegaDepth [17], HRWSI [19], or a mix of multiple datasets [16] [26] [27]. In recent years, domain adaptation was also applied to this problem [46] [47]. We find that it is essentially due to not only the diversity of scene semantics, but more importantly, the diversity of scene scales that limits the generalization ability of depth inference in unseen scenes. In the literature, scales and semantics of output scenes are jointly considered in the sub-task of depth estimation [7] [17] [19] such as famous MiDas [16], which normalized scene scales to generate relative depth maps. However, the generalization issue of monocular depth inference remains a challenge in the sub-tasks of depth completion [21] [39] and depth enhancement [13] [41], which require to generate absolute depth maps.

# 3 G2-MONODEPTH LOSS

## 3.1 Problem Definition

We suppose a robot equipped with unknown configuration of sensors $\theta$ to perceive an unknown scene $\eta$. G2-MonoDepth aims to infer a high-quality depth map $d$ from a single RGB image $I(\eta)$ plus an optional raw depth map X. X map can be considered as a depth map generated from a GT depth map $z(\eta)$ by a degradation function $x(\cdot)$, denoted as $X = x(z(\eta), \theta)$. Therefore, the task of G2-MonoDepth can be formulated as follows:

$$d = f(I(\eta), x(z(\eta), \theta)), \tag{1}$$

where $f(\cdot)$ is a mapping function. RGB image $I(\eta)$ and GT depth map $z(\eta)$ may have diverse scales and semantics for different scenes $\eta$, such as indoor bedroom and outdoor traffic. Similarly, X map $x(z(\eta), \theta)$ may have arbitrary [0%, 100%] valid depth pixels with diverse depth errors for different sensors $\theta$, such as ToF, SL, or 4/16/64/128-line LiDAR. For simplicity, the notations of scene $\eta$ and sensor $\theta$ are omitted in the rest of the paper. Therefore, the notations $d(\eta)$, $I(\eta)$, $z(\eta)$, and $x(z(\eta), \theta)$ are simply denoted as $d$, $I$, $z$, and $x(z)$, respectively.

## 3.2 Formulation of G2-MonoDepth Loss

Our G2-MonoDepth loss aims to establish a constraint between inferred depth map $d$ and GT depth map $z$ conditioned upon RGB image $I$ and X map $x(z)$. Specifically, our G2-MonoDepth loss considers both relative relation and absolute relation to accommodate diverse sparsity/errors of input raw data and diverse scales of unseen output scenes. We investigate the effect of the two conditions $I$ and $x(z)$ on the constraint, separately.

We first consider the relation between inferred depth map $d$ and GT depth map $z$ conditioned upon a single RGB image $I$. It is acknowledged that the scale of inferred depth map $d$ is unknown because of the scale-ambiguity problem [7]. However, the relative relation of objects does not vary with the scales of the scene. It is reasonable to consider a relative relation between inferred depth map $d$ and GT depth map $z$ rather than an absolute relation. Therefore, their relation can be approximately formulated with a basic expression [16] as follows:

$$sd_i + f = z_i, s \in R^+, f \in R, \tag{2}$$

where $s$ is a scale factor, and $f$ is a shift factor. $d_i$ and $z_i$ denote the values at the location $i \in \{1, 2, \ldots, M\}$ in inferred depth map $d$ and GT depth map $z$, respectively. $M$ denotes the number of pixels in GT depth map.

Then, we consider the relation between inferred depth map $d$ and GT depth map $z$ conditioned upon a single X map $x(z)$. In our task, X map $x(z)$ only provides the intensities of valid depth pixels (i.e., arbitrary [0%, 100%] valid pixels), whereas the intensities of other depth pixels are unknown. Therefore, it is reasonable to only consider an absolute relation of these valid depth pixels between

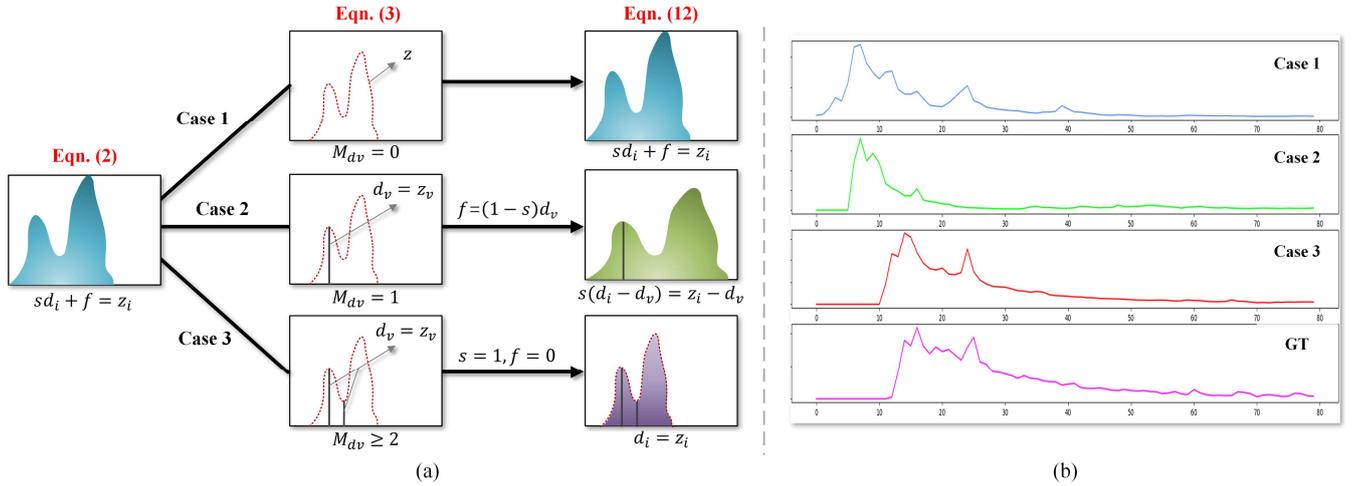

Fig. 2. Illustrate the decomposition from the two relations in Eqns. (2) and (3) to Eqn. (12). (a) Three cases of regressions, and (b) an example of inferred depth map from KITTI. The figures show the histograms of inferred depth maps and GT. In Case 1, both the scale $s$ and shift $f$ of inferred depth map are ambiguous compared with GT. In Case 2, only the scale $s$ of inferred depth map is ambiguous compared with GT. In Case 3, both the scale $s$ and shift $f$ of inferred depth map are the same as GT.

inferred depth map $d$ and GT depth map $z$ as follows:

$$d_v = z_v, v \in V, \quad (3)$$

where $V \subseteq \{1, 2, \ldots, M\}$ denotes the locations of all valid depth pixels in X map. $d_v$ and $z_v$ denote the values at the location $v$ in inferred depth map $d$ and GT depth map $z$, respectively. We denote the size of the location set $V$ to be $M_V$. Because X map may have arbitrary [0%, 100%] valid pixels, we have $M_V \leq M$.

These two relations in Eqns. (2) and (3) jointly establish a constraint between inferred depth map $d$ and GT depth map $z$ for G2-MonoDepth given monocular RGB+X data. However, the two factors $s$ and $f$ in the relation in Eqn. (2) are unknown. We equivalently derive a computable relation from Eqn. (2).

By taking the mean of Eqn. (2), we obtain the relation of two mean values $\bar{d}$ and $\bar{z}$ as follows:

$$\bar{z} = \frac{1}{M} \sum_{i=1}^{M} z_i = \frac{1}{M} \sum_{i=1}^{M} (sd_i + f) = s\bar{d} + f. \quad (4)$$

Similarly, by taking the mean derivation of Eqn. (2), we obtain the relation of two mean deviation values $\sigma_d$ and $\sigma_z$ as follows:

$$\sigma_z = \frac{1}{M} \sum_{i=1}^{M} |z_i - \bar{z}| = \frac{1}{M} \sum_{i=1}^{M} |sd_i + f - s\bar{d} - f| = s\sigma_d. \quad (5)$$

By standardizing Eqn. (2) with $\bar{z}$ and $\sigma_z$, and then substituting (4) and (5) into it, we yield

$$\frac{z_i - \bar{z}}{\sigma_z + \epsilon} = \frac{sd_i + f - \bar{z}}{\sigma_z + \epsilon} = \frac{sd_i + f - (s\bar{d} + f)}{s\sigma_d + \epsilon} = \frac{d_i - \bar{d}}{\sigma_d + \frac{\epsilon}{s}}, \quad (6)$$

where $\epsilon$ ($0 < \epsilon \ll \sigma_z$) is a factor to prevent zero denominator.

Considering that the factor $\epsilon$ is very small, we approximately have

$$\frac{d_i - \bar{d}}{\sigma_d + \epsilon} \approx \frac{z_i - \bar{z}}{\sigma_z + \epsilon}. \quad (7)$$

Eqns. (4)–(7) prove that the relative relation in Eqn. (2) is approximately equivalent to Eqn. (7). However, the two unknown factors, $s$ and $f$, have been eliminated in Eqn.

(7). Then, a scale-adaptive loss $\mathcal{L}_{sa}(d, z)$ is formulated on the basis of the relations (3) and (7) as follows:

$$\mathcal{L}_{sa}(d, z) = \frac{1}{M} \sum_{i=1}^{M} \left| \frac{d_i - \bar{d}}{\sigma_d + \epsilon} - \frac{z_i - \bar{z}}{\sigma_z + \epsilon} \right| + \frac{1}{M_V + \epsilon} \sum_{v=1}^{M_V} |d_v - z_v| \quad (8)$$

The factor $\epsilon$ aims to prevent zero denominator, because the number of valid depth pixels $M_V$ could be zero in our task. Notably, we adopt the mean deviation in Eqn. (5) instead of the standard deviation ($\sqrt{1/M \sum_{i=1}^{M}(x_i - \bar{x})^2}$) in z-score standardization to improve training stability. The ablation study on this operation is shown in Section 5.

Besides, we additionally adopt the multi-scale scale-invariant gradient term [16] [17] [19] to regularize inferred depth maps as follows:

$$\mathcal{L}_{sg}(d, z) = \frac{1}{M} \sum_{k=1}^{4} \sum_{i=1}^{M} \left( |\nabla_h R_i^k| + |\nabla_w R_i^k| \right), \quad (9)$$

where

$$R_i^k = \rho_k \left( \frac{d_i - \bar{d}}{\sigma_d + \epsilon} - \frac{z_i - \bar{z}}{\sigma_z + \epsilon} \right). \quad (10)$$

$\nabla_h R_i^k$ and $\nabla_w R_i^k$ are the gradients of $R_i^k$ in the $h$ and $w$ directions, respectively. $\rho_k(\cdot)$ is the down-sampling function with the scale $k$ ($1 \leq k \leq 4$). Notably, the traditional differential operator in [16] [17] [19] is replaced by a Sobel operator in (9) to further improve the quality of inferred depth map. The ablation study on this operation is shown in Section 5.

Finally, our G2-MonoDepth loss is formulated as follows:

$$\mathcal{L}_{G2-MonoDepth}(d, z) = \mathcal{L}_{sa}(d, z) + \lambda \mathcal{L}_{sg}(d, z), \quad (11)$$

where $\lambda$ is the regularization coefficient being set as 0.5 empirically [16] [17].

### 3.3 Relations to Other Losses

Our G2-MonoDepth loss is derived from the basic relations (2) and (3). Eqn. (2) can be regarded as one or multiple binary linear equation(s) with unknown variables $s$ and $f$ by substituting (3) into it. We denote the number of equations in Eqn. (3) as $M_{dv}$, which is determined by the number of possible intensities $d_v$ of valid depth pixels in X map. $M_{dv}$ is generally smaller than the number of all valid depth pixels $M_v$, because multiple valid depth pixels may have the same intensity value.

Thus, we equivalently decompose the relations (2) and (3) into three cases of regressions in (12) according to the value of $M_{dv}$.

$$\begin{cases} sd_i + f = z_i, s \in R^+, f \in R, & M_{dv} = 0 \\ s(d_i - d_v) = z_i - d_v, s \in R^+, & M_{dv} = 1 \\ d_i = z_i, & M_{dv} \geq 2 \end{cases} \quad (12)$$

The three cases of regressions in (12) are elaborated as follows.

**Case 1: $M_{dv} = 0$.** In this case, the relations (2) and (3) are equivalent to a single relation (2), that is, $sd_i + f = z_i, s \in R^+, f \in R$. It is an *affine-invariant regression* [16] with unknown global scale $s$ and global shift $f$.

**Case 2: $M_{dv} = 1$.** By substituting a single equation of $d_v = z_v$ in Eqn. (3) into Eqn. (2), the shift $f$ could be solved as $f = (1-s)d_v$. Then, the relations (2) and (3) are equivalent to $s(d_i - d_v) = z_i - d_v, s \in R^+$. It belongs to a *scale-invariant regression* [7] [17] with unknown global scale $s$.

**Case 3: $M_{dv} \geq 2$.** By substituting two or more equations of $d_v = z_v$ in Eqn. (3) into Eqn. (2), the scale $s$ and the shift $f$ could be solved as $s = 1$ and $f = 0$. In this case, the relations (2) and (3) are equivalent to $d_i = z_i$. Hence, it is a *direct regression* [4] [12] [20], which is widely used in most models.

The decomposition of the relations (2) and (3) is further illustrated in Fig. 2. Fig. 2(a) shows three cases of regressions by the histograms of inferred depth map $d$. In Case 1, both the scale $s$ and shift $f$ of the histogram are unfixed, indicating that inferred depth map $d$ is scale-ambiguous and shift-ambiguous compared with GT depth map $z$. In Case 2, only the scale $s$ of the histogram is unfixed, indicating that inferred depth map $d$ is scale-ambiguous compared with GT depth map $z$. In Case 3, the scale $s$ and the shift $f$ of the histogram are fixed, indicating that inferred depth map $d$ is the same as GT depth map $z$. Fig. 2(b) provides a practical example of inferred depth map in KITTI. The scale and shift of the histogram of inferred depth map are different from GT depth map in Case 1. The histogram of inferred depth map is partly rectified in Case 2. In Case 3, the histogram of inferred depth map is almost the same as GT depth map.

Then, we analyze the relationship of our G2-MonoDepth loss in Eqn. (11) with other representative losses of monocular depth inference based on Eqn. (12).

(a) **L1 and L2 losses.** These two losses are widely used in multiple sub-tasks of monocular depth inference, such as depth estimation [10], depth completion [4], and depth enhancement [5]. These losses impose an absolute relation between inferred depth map $d$ and GT depth map $z$ as follows:

$$d_i = z_i . \quad (13)$$

Therefore, L1 and L2 losses are generally formulated as follows:

$$\mathcal{L}_1 = \frac{1}{M} \sum_{i=1}^{M} |d_i - z_i|, \mathcal{L}_2 = \frac{1}{M} \sum_{i=1}^{M} (d_i - z_i)^2 . \quad (14)$$

The absolute relation of L1 and L2 losses in Eqn. (13) corresponds to **Case 3** in Eqn. (12).

(b) **Scale-invariant loss** [7]. This loss usually infers relative depth maps with unknown global scales. It normalizes the global scale between inferred depth map $d$ and GT depth map $z$ as follows:

$$sd_i = z_i, s \in R^+. \quad (15)$$

This equation can be transformed into a computable loss function with unknown $s$ in log-space as follows:

$$\mathcal{L}_{si}(d, z) = \frac{1}{M} \sum_{i=1}^{M} \Delta^2 - \frac{1}{M} \left( \sum_{i=1}^{M} \Delta \right)^2, \quad (16)$$

where $\Delta = \log d_i - \log z_i$.

The constraint of scale-invariant loss in Eqn. (15) corresponds to **Case 2** in Eqn. (12).

(c) **Affine-invariant loss** [16]. This loss usually infers relative depth maps with unknown global scale $s$ and unknown global shift $f$. It normalizes both the global scale and global shift between inferred depth map $d$ and GT depth map $z$ similar to Eqn. (2). A computable loss is then derived on the basis of Eqn. (2) by estimating a global scale and a global shift as follows:

$$\mathcal{L}_{ai}(d, z) = \frac{1}{M} \sum_{i=1}^{M} \left| \frac{d_i - f(d)}{s(d)} - \frac{z_i - f(z)}{s(z)} \right|, \quad (17)$$

where $f(\cdot)$ and $s(\cdot) = 1/M \sum_{i=1}^{M} |d_i - f(d)|$ denote the median and median deviation, respectively.

The constraint of affine-invariant loss in Eqn. (2) corresponds to **Case 1** in Eqn. (12).

(d) **Ranking loss** [6]. This loss infers relative depth maps with unknown local scales. It normalizes the local scales of all pixels between inferred depth map $d$ and GT depth map $z$ as follows:

$$s_i d_i = z_i, s_i \in R^+, \quad (18)$$

where $s_i$ denotes the local scale of the pixel at the location $i$. A computable loss that meets the constraint (18) was given by

$$\mathcal{L}_{rk}(d_i, d_j) = \begin{cases} \log \left( 1 + e^{-l_{ij}(d_i - d_j)} \right), & l_{ij} \neq 0 \\ (d_i - d_j)^2, & l_{ij} = 0 \end{cases}, \quad (19)$$

where $l_{ij} \in \{-1, 0, 1\}$ is the ordinal label of the two pixels at the locations $i, j$ in GT depth map.

The constraint of ranking loss in Eqn. (18) degrades to **Case 2** in Eqn. (12) when the local scales of all pixels $s_i$ are equal.

# 4 NETWORK AND TRAINING DATASET

## 4.1 ReZero U-Net

In our solution, the network architecture requires to learn a mapping from the input of RGB+X data to the output of high-quality depth maps. According to our G2-MonoDepth loss in (11), the relative relation of inferred depth maps is mainly estimated from the input of RGB image while the absolute scale of inferred depth map is mainly propagated from the input of X. Most networks of monocular depth inference could well estimate the relative relation in unseen scenes such as [7] [16] [17]. However, we find that the widely-used Batch Normalization (BN) [18] in various network architectures may hinder the scale propagation from the input of X maps to the output of inferred depth maps on training datasets with diverse scene scales, though it has achieved a great success in CNNs for many tasks such as image classification [22] and depth inference [3] [7] [16]. Directly using these BN-based networks may limit the generalization ability of our unified model in unseen scenes with diverse scales.

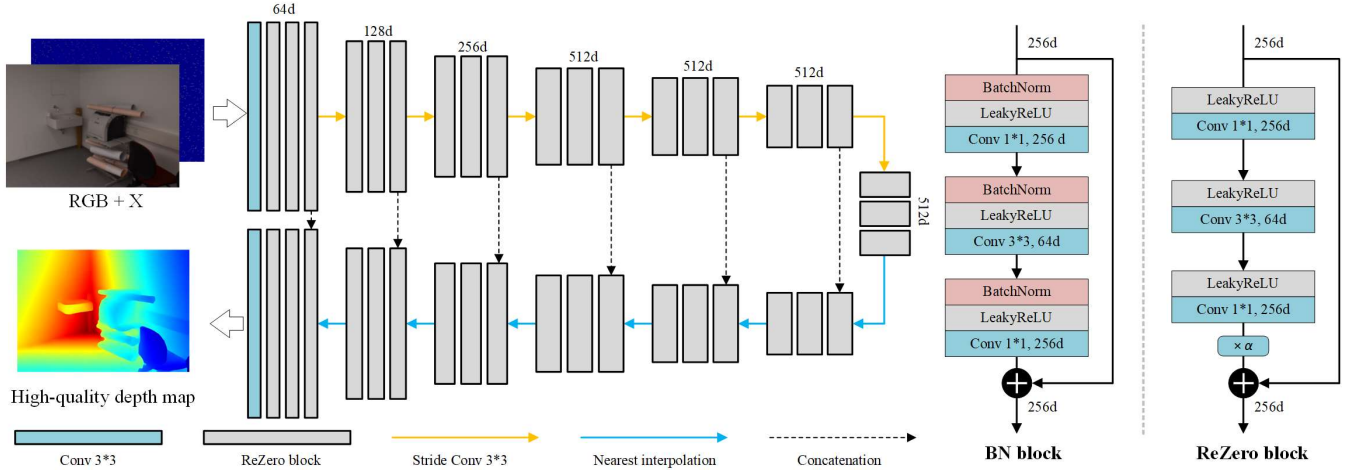

Fig. 3. Architecture of ReZero U-Net. A basic U-Net is adopted for G2-MonoDepth; however, the involved BN is replaced by ReZero for scale propagation from the input of X maps to the output of inferred depth maps on training datasets with diverse scene scales.

BN is generally expressed as

$$b = \frac{a - \bar{a}}{\sigma'_a} * \gamma + \eta \,, \qquad (20)$$

where $a, b$ denote the mini-batch data of input and output in BN, respectively; $\bar{a}, \sigma'_a$ denote mean and standard deviation of input $a$, respectively; and $\gamma, \eta$ are learnable parameters. When the input $a$ is scaled to $sa + f$ by an unknown scale factor $s$ and shift factor $f$, the output $b$ can be scaled proportionally to $sb + f$ only if $\gamma = s$ and $\eta = f$. However, $s$ and $f$ would significantly change in different mini-batches because of the diversity of scene scales/semantics in our task. Therefore, BN layers may hinder the propagation of scale and shift from X maps to inferred depth maps in the training. Similar problems were discussed in image super-resolution [23] and deblurring [24].

Therefore, we fully remove BN for the scale propagation and further add ReZero [25] for stable training in the basic blocks of the network architecture. The used ReZero block in our network is shown in Fig. 3, which is modified from the BN block in [61] and the learnable parameter is initialized to be $\alpha = 0$. It ensures that our network well propagates absolute scene scales from input to output. The ablation study on BN and ReZero blocks is shown in Section 5. BN block could degrade the generalization ability of our model in unseen scenes with different scales.

Notably, we adopt a basic U-Net [60] with low complexity to establish a basic benchmark for our unified task, rather than other complicated architectures such as DenseNet [49], HourglassNet [7], or ViTs [44] [45]. Our network architecture is shown in Fig. 3. Two 3×3 convolutions are used in the first and last layers to adjust the channel numbers of feature maps. Spatial resolution of feature maps is halved by stride convolution with stride 2 in the encoder and doubled by the nearest interpolation in the decoder. Skip connections are achieved by concatenating the feature maps of the encoder to the decoder.

## 4.2 Augmented Training Dataset

The training of our G2-MonoDepth requires a large-scale RGB-D dataset with diverse scene scales/semantics and depth sparsity/errors, which includes RGB images, GT depth maps, and raw depth maps. Though a large quantity of public RGB-D datasets has been released in recent years, most of them do not provide raw depth maps with GT labels [65]. Though a few datasets provide raw depth maps with GT labels such as NYUv2 [14] and KITTI [15], sparsity/errors of raw depth maps are not diverse enough. It is hard to collect a large-scale RGB-D dataset with diverse scene scales/semantics and depth sparsity/errors, which includes RGB images, GT depth maps, and raw depth maps simultaneously. Therefore, using artificially augmented data [4] [13] [51] [58] or GANs [8] [42] has been the dominant solution to train models in relevant tasks of RGB-D data in the literature.

In this paper, we develop a data augmentation pipeline in Fig. 4 to well simulate all types of real artifacts in raw depth maps from all kinds of depth sensors in practical scenarios including depth sparsity/errors [5] [13] [41]. First, Salt&Pepper noises with random probabilities and Gaussian noises with random standard deviations are imposed on GT depth maps. Second, blurs are generated by down-sampling and up-sampling GT depth maps with different zoom factors (i.e., 2, 4, 8, and 16) [20] [40] [56]. Third, sparse depth pixels are sampled from GT depth maps with random sampling rates [4] [12] [55]. Finally, holes are collected from multiple RGB-D datasets, and then augmented by random cropping, affine transform, and flipping operations, and then randomly imposed to GT depth maps.

Our data augmentation pipeline provides an optional way to generate a training dataset with diverse depth sparsity/errors and scene scales/semantics for our unified model. Specifically, our augmented training dataset is generated from three public datasets Matterport3D [32],

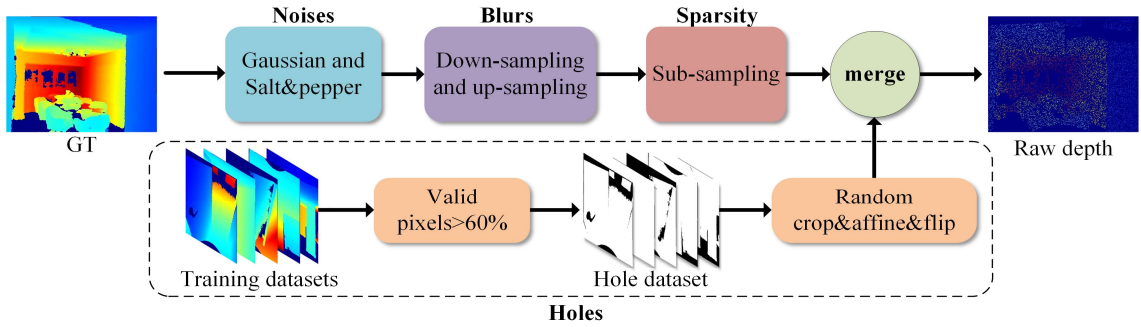

Fig. 4. Data augmentation pipeline to generate our training dataset with diverse depth sparsity/errors and scene scales/semantics from mixed RGB-D datasets.

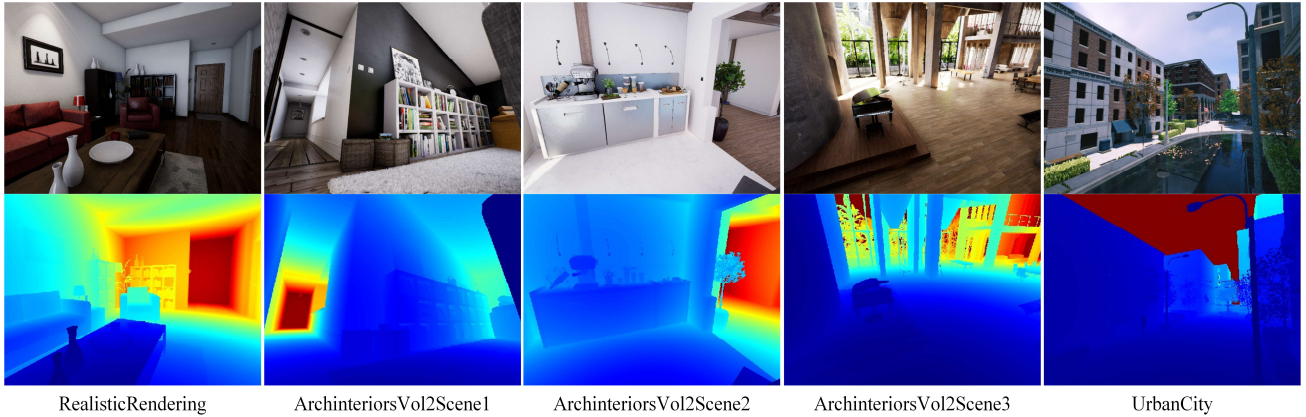

RealisticRendering        ArchinteriorsVol2Scene1        ArchinteriorsVol2Scene2        ArchinteriorsVol2Scene3        UrbanCity

Fig. 5. A new dataset UnrealCV collected by the UnrealCV project [28].

HRWSI [19], vKITTI [33], and a new dataset called UnrealCV in Table 1 based on the developed pipeline. These datasets contain various indoor and outdoor scenes as well as real and synthetic scenes. The new dataset, UnrealCV, is collected by the UnrealCV project [28] on Unreal Engine 4. This dataset provides 5000 realistic RGB images and high-quality depth maps from four indoor scenes and one outdoor scene, as shown in Fig. 5. Our augmented training dataset could well improve the generalization ability of our unified model in real scenes. In addition, it can be used in the individual sub-tasks of monocular depth inference in real scenes by other researchers such as depth estimation, depth completion with different sparsity, and depth enhancement.

## 5 EXPERIMENTS

### 5.1 Experiment Setting

**Training.** Our G2-MonoDepth is trained on the augmented training dataset in Section 4.2, which is generated from three public datasets (Matterport3D [32], HRWSI [19], and vKITTI [33]) and a new dataset (UnrealCV). The network is randomly initialized by He et al. [62] with normal distribution. We adopt AdamW as the optimizer with a learning rate of $2 \times 10^{-4}$ and $(\beta_1 = 0.9, \beta_2 = 0.999)$ using two RTX 3090, and adopt cosine learning rate decay with 100 epochs. The batch size is set as 64 (32/GPU). The mixed precision learning is adopted to accelerate the training by the Automatic Mixed Precision (AMP) of PyTorch. Training data are flipped horizontally with 50% probability. In the data augmentation pipeline, the standard deviation of Gaussian noises ranges [0.01, 0.1], the probability of Salt&Pepper noises ranges [0, 1], the zoom factor of down-sampling and up-sampling is selected from the set {2, 4, 8, 16}, and the percentage of valid depth pixels in raw depth maps ranges [0%, 100%]. All depth maps in the training dataset are uniformly rescaled to [0, 1], and their resolutions are resized to 320 in height.

TABLE 1. TRAINING AND TEST DATASETS OF G2-MONODEPTH

|       | Dataset | Scene | Size | Type |
|-------|---------|-------|------|------|
| **Train** | Matterport3D [32] | Indoor | 194K | SL |
|       | HRWSI [19] | Outdoor | 20K | Multiview |
|       | vKITTI [33] | Outdoor | 21K | Synthetic |
|       | UnrealCV | Indoor/outdoor | 5K | Synthetic |
| **Test** | Ibims [34] | Indoor | 100 | LiDAR |
|       | KITTI [15] | Outdoor | 1000 | Multiview |
|       | NYUv2 [14] | Indoor | 654 | SL |
|       | DIODE [35] | Indoor/outdoor | 771 | LiDAR |
|       | ETH3D [36] | Indoor/outdoor | 454 | LiDAR |
|       | Sintel [37] | Indoor/outdoor | 1064 | Synthetic |

**Testing.** Our G2-MonoDepth is applied in the three individual sub-tasks of depth estimation, depth completion with different sparsity, and depth enhancement in unseen scenes. First, our model is comprehensively evaluated on seven real-world datasets that are not seen in the training including Ibims [34], KITTI [15], NYUv2 [14], DIODE [35], ETH3D [36], and Sintel [37] in Table 1 as well as RealSense [2]. In these real-world datasets, scene scales, scene semantics, depth sparsity (4/8/16/32/64-line), and depth errors (e.g., noises, blurs, holes) vary drastically.

TABLE 2. QUANTITATIVE RESULTS OF G2-MONODEPTH AND THE BASELINES ON REAL-WORLD DATA

| Methods | Ibims | | NYUv2 | | KITTI | | DIODE | | ETH3D | | Sintel | | Rank ↓ | |
|---|---|---|---|---|---|---|---|---|---|---|---|---|---|---|
| | OE | SRMSE | OE | SRMSE | OE | SRMSE | OE | SRMSE | OE | SRMSE | OE | SRMSE | OE | SRMSE |
| DMF [38] | 0.249 | 0.88 | 0.205 | 0.74 | 0.186 | 0.95 | 0.334 | 1.32 | 0.240 | 0.91 | 0.343 | 1.35 | 4.5 | 4.0 |
| SGRL [19] | 0.234 | 0.83 | 0.202 | 0.73 | **0.110** | <u>0.89</u> | 0.274 | 1.17 | 0.200 | 0.83 | <u>0.290</u> | <u>1.18</u> | 2.5 | 2.7 |
| Mega [17] | 0.300 | 1.11 | 0.242 | 0.93 | 0.142 | **0.83** | 0.330 | 1.35 | 0.228 | 0.87 | 0.351 | 1.38 | 4.3 | 4.2 |
| MiDaS [16] | <u>0.165</u> | <u>0.69</u> | <u>0.158</u> | <u>0.65</u> | <u>0.121</u> | 0.96 | <u>0.210</u> | <u>1.02</u> | <u>0.154</u> | <u>0.77</u> | **0.276** | **1.17** | <u>1.8</u> | <u>2.3</u> |
| Ours | **0.145** | **0.56** | **0.157** | **0.61** | 0.156 | | **0.208** | **1.00** | **0.150** | **0.66** | 0.310 | 1.26 | **1.8** | **1.7** |

| Methods | NYUv2 | | KITTI | | | | | | | | | | Rank ↓ | |
|---|---|---|---|---|---|---|---|---|---|---|---|---|---|---|
| | | | 64-lines | | 32-lines | | 16-lines | | 8-lines | | 4-lines | | | |
| | Abs | RMSE | Abs | RMSE | Abs | RMSE | Abs | RMSE | Abs | RMSE | Abs | RMSE | Abs | RMSE |
| SDCM [11] | 0.467 | 33.79 | 0.048 | 5.92 | 0.067 | 7.31 | 0.113 | 11.35 | 0.199 | 15.08 | 0.336 | 27.84 | 4.2 | 3.7 |
| GuideNet [39] | 1.580 | 97.28 | <u>0.022</u> | **3.17** | 0.054 | <u>5.26</u> | 0.115 | 8.76 | 0.275 | 16.87 | 0.567 | 28.18 | 4.3 | 3.2 |
| TWISE [21] | 0.084 | 10.97 | **0.019** | 6.55 | <u>0.034</u> | 10.39 | 0.066 | 16.32 | 0.134 | 21.27 | 0.372 | 41.35 | 2.7 | 4.7 |
| NLSPN [12] | <u>0.025</u> | <u>5.49</u> | 0.035 | 5.61 | 0.038 | 6.33 | <u>0.046</u> | <u>7.67</u> | <u>0.077</u> | <u>10.54</u> | <u>0.193</u> | <u>16.64</u> | <u>2.5</u> | <u>2.3</u> |
| Ours | **0.015** | **3.69** | 0.025 | <u>4.23</u> | **0.028** | **4.76** | **0.035** | **5.88** | **0.045** | **7.25** | **0.070** | **9.88** | **1.3** | **1.2** |

| Methods | Ibims | | NYUv2 | | KITTI | | DIODE | | ETH3D | | Sintel | | Rank ↓ | |
|---|---|---|---|---|---|---|---|---|---|---|---|---|---|---|
| | Abs | RMSE | Abs | RMSE | Abs | RMSE | Abs | RMSE | Abs | RMSE | Abs | RMSE | Abs | RMSE |
| DKN [13] | 0.294 | 9.83 | 0.120 | 10.50 | 0.637 | 34.93 | 0.365 | 14.96 | 0.783 | 15.43 | 1.729 | 9.61 | 4.3 | 4.2 |
| PMBAN [5] | 0.166 | 6.21 | 0.069 | 6.55 | 0.652 | 36.45 | 0.235 | 11.78 | 0.809 | 15.62 | <u>0.969</u> | 6.42 | 4.0 | 4.2 |
| PDSR [40] | 0.135 | 5.26 | 0.055 | 5.36 | 0.642 | 35.50 | 0.221 | 11.27 | 0.787 | 15.59 | 1.042 | 15.09 | 3.3 | 3.7 |
| JIFF [41] | <u>0.068</u> | <u>3.04</u> | <u>0.028</u> | **3.06** | <u>0.599</u> | <u>31.55</u> | <u>0.158</u> | <u>8.66</u> | <u>0.705</u> | <u>13.94</u> | 1.203 | <u>5.56</u> | <u>2.3</u> | <u>1.8</u> |
| Ours | **0.027** | **2.11** | **0.024** | <u>3.87</u> | **0.067** | **5.78** | **0.139** | **7.40** | **0.109** | **2.16** | **0.459** | **5.40** | **1.0** | **1.2** |

*The table shows the quantitative results on the six test datasets in the three sub-tasks of depth estimation (top), depth completion with different sparsity (center), and depth enhancement (bottom). "Rank" [19] denotes the average of rank numbers (i.e., 1, 2, and 3) over all the test datasets. The best results are in **bold** and the 2nd-best results are <u>underlined</u>. Notably, the four datasets Ibims, DIODE, ETH3D, and Sintel do not provide real raw depth data for depth completion.*

Second, our model is additionally evaluated on synthetic datasets generated from the six datasets Ibims, KITTI, NYUv2, DIODE, ETH3D, and Sintel by our data augmentation pipeline. In these synthetic datasets, the sparsity (0%, 0.1%, 1%, 10%, and 100%) and errors are more diverse than the real-world datasets. Notably, GT depth maps in these test datasets may be sparse or contain holes such as Ibims, KITTI, DIODE, and ETH3D. Thus, 100% refers to all valid pixels in GT depth maps in the test.

**Baselines.** A total of twelve SOTA baselines in recent years are used for comparison including: (1) Mega [17], SGRL [19], MiDaS [16], and DMF [38] in the sub-task of depth estimation; (2) SDCM[2] [11], NLSPN [12], GuideNet [39], and TWISE [21] in the sub-task of depth completion; (3) PDSR [40], PMBAN [5], DKN [13], and JIFF [41] in the sub-task of depth enhancement. These baselines well addressed one sub-tasks of monocular depth inference and made great advances in public datasets. By comparison, our G2-MonoDepth handles all these sub-tasks with a unified model. The well-trained models released by the authors are directly adopted to obtain desired results on the test datasets. Notably, the baselines NLSPN, PMBAN, DKN, and JIFF were trained in the training sets of NYUv2 (our test dataset), and the baselines SDCM, GuideNet, and TWISE were trained in the training sets of KITTI (our test dataset). By comparison, our G2-MonoDepth is always tested in unseen datasets. All test data are adjusted according to the experiment setting of the baselines, such as the range and resolution of depth maps.

**Metrics.** Four widely-used metrics are utilized to measure the quantitative results of inferred depth maps,

including ordinal error (OE) [6] [19], standardized root mean squared error (SRMSE) [16], root mean squared error (RMSE), and absolute relative error (Abs). The expressions of these metrics are as follows:

(a) OE: $\sum_i \mathbb{1}\left(o_{ij}(d) \neq o_{ij}(z)\right)$, $o_{ij}(d) = \begin{cases} +1, & \frac{d_i}{d_j} \geq 1 + \tau \\ -1, & \frac{d_i}{d_j} \leq \frac{1}{1+\tau} \\ 0, & \text{else} \end{cases}$,

(b) SRMSE: $\sqrt{\frac{1}{n}\sum_{i=1}^n \left(\frac{d_i - \bar{d}}{\sigma_d + \epsilon} - \frac{z_i - \bar{z}}{\sigma_z + \epsilon}\right)^2}$,

(c) RMSE: $\sqrt{\frac{1}{n}\sum_{i=1}^n (d_i - z_i)^2}$,

(d) Abs: $\frac{1}{n}\sum_{i=1}^n \frac{|d_i - z_i|}{z_i}$,

where $\tau$ is a tolerated threshold set as 0.01. Notably, the metrics OE and SRMSE are only used for the sub-task of depth estimation, which measure the relative relation between inferred depth maps and GT depth maps. The metrics RMSE and Abs are used for the other sub-tasks of depth completion and depth enhancement, which measure the absolute relation between inferred depth maps and GT depth maps.

## 5.2 Comparison and Analysis

**Quantitative results on real-world data.** Table 2 shows the quantitative results of our G2-MonoDepth and the twelve baselines on the real-world data in the three sub-tasks of depth estimation, depth completion with different sparsity, and depth enhancement. Our G2-MonoDepth always achieves the best results in all the scenarios on all the metrics compared with the baseline models. In the sub-task of depth estimation, our G2-MonoDepth performs a little better than the famous MiDas and much better than the other baselines on

---

[2] SDCM [11] released both an unsupervised model and a supervised one in *https://github.com/fangchangma/self-supervised-depth-completion*. We adopt the latter one as the baseline.

TABLE 3. QUANTITATIVE RESULTS OF G2-MONODEPTH AND THE BASELINES ON SYNTHETIC DATA

| Sparsity | Methods | Ibims | | NYUv2 | | KITTI | | DIODE | | ETH3D | | Sintel | | Rank | |
|---|---|---|---|---|---|---|---|---|---|---|---|---|---|---|---|
| | | Abs | RMSE | Abs | RMSE | Abs | RMSE | Abs | RMSE | Abs | RMSE | Abs | RMSE | Abs | RMSE |
| 0.1% | SDCM [11] | 0.287 | 10.97 | 0.335 | 25.98 | 0.329 | 26.53 | 0.422 | 19.14 | 1.497 | 14.50 | 10.276 | 25.44 | 2.3 | 2.7 |
| | GuideNet [39] | 0.521 | 15.83 | 0.350 | 32.67 | 0.713 | 31.83 | 1.145 | 29.75 | 2.093 | 22.63 | 13.096 | 42.08 | 4.5 | 4.5 |
| | TWISE [21] | 0.290 | 11.44 | 0.268 | 26.67 | 0.577 | 40.69 | 0.594 | 22.65 | 2.165 | 24.89 | 13.321 | 33.14 | 3.8 | 4.2 |
| | NLSPN [12] | 0.291 | 9.05 | 0.127 | 12.96 | 0.409 | 21.13 | 0.515 | 15.35 | 2.874 | 25.62 | 11.498 | 25.50 | 3.3 | 2.7 |
| | Ours | 0.078 | 3.40 | 0.051 | 6.02 | 0.151 | 12.03 | 0.266 | 10.10 | 0.912 | 9.84 | 3.193 | 10.63 | 1.0 | 1.0 |
| 1% | SDCM [11] | 0.164 | 6.98 | 0.106 | 11.29 | 0.183 | 14.93 | 0.287 | 13.14 | 0.835 | 9.09 | 7.060 | 17.18 | 3.0 | 3.8 |
| | GuideNet [39] | 0.235 | 6.84 | 0.114 | 11.07 | 0.331 | 16.90 | 0.431 | 12.31 | 1.588 | 16.72 | 11.996 | 30.10 | 5.0 | 4.3 |
| | TWISE [21] | 0.156 | 5.67 | 0.069 | 7.72 | 0.174 | 19.66 | 0.364 | 11.72 | 1.203 | 13.19 | 7.924 | 18.09 | 3.0 | 3.5 |
| | NLSPN [12] | 0.177 | 6.43 | 0.079 | 7.59 | 0.179 | 12.16 | 0.315 | 11.32 | 1.069 | 11.21 | 5.191 | 12.72 | 3.0 | 2.3 |
| | Ours | 0.035 | 2.28 | 0.029 | 4.14 | 0.066 | 7.04 | 0.173 | 8.10 | 0.262 | 3.47 | 1.060 | 6.14 | 1.0 | 1.0 |
| 10% | SDCM [11] | 0.288 | 9.81 | 0.199 | 16.08 | 0.217 | 15.19 | 0.341 | 16.24 | 0.496 | 6.80 | 4.299 | 13.45 | 4.2 | 4.2 |
| | GuideNet [39] | 0.143 | 5.27 | 0.098 | 8.78 | 0.176 | 10.00 | 0.271 | 9.86 | 1.083 | 10.82 | 8.360 | 20.69 | 3.8 | 3.5 |
| | TWISE [21] | 0.130 | 5.43 | 0.068 | 7.16 | 0.142 | 14.51 | 0.272 | 10.71 | 0.755 | 9.09 | 6.043 | 14.70 | 2.8 | 3.3 |
| | NLSPN [12] | 0.208 | 7.30 | 0.086 | 8.05 | 0.156 | 9.23 | 0.321 | 11.12 | 0.534 | 7.47 | 4.086 | 11.14 | 3.2 | 3.0 |
| | Ours | 0.028 | 2.09 | 0.025 | 3.86 | 0.055 | 5.79 | 0.146 | 7.46 | 0.104 | 2.13 | 0.506 | 5.32 | 1.0 | 1.0 |

*The table shows the quantitative results on synthetic data (with the sparsity 0.1%, 1%, and 10%) from the six test datasets in the sub-task of depth completion.*

TABLE 4. QUANTITATIVE RESULTS OF G2-MONODEPTH AND THE BASELINES ON REAL-WORLD DATA (RETRAINED)

| Methods | Ibims | | NYUv2 | | KITTI | | DIODE | | ETH3D | | Sintel | | Rank ↓ | |
|---|---|---|---|---|---|---|---|---|---|---|---|---|---|---|
| | OE | SRMSE | OE | SRMSE | OE | SRMSE | OE | SRMSE | OE | SRMSE | OE | SRMSE | OE | SRMSE |
| MiDas# | 0.219 | 0.75 | 0.294 | 0.94 | 0.220 | 1.16 | 0.284 | 1.18 | 0.222 | 0.85 | 0.345 | 1.33 | 2.0 | 1.8 |
| Ours# | 0.161 | 0.62 | 0.163 | 0.61 | 0.156 | 0.96 | 0.218 | 1.02 | 0.165 | 0.70 | 0.341 | 1.34 | 1.0 | 1.2 |

| Methods | NYUv2 | | KITTI | | | | | | | | | | Rank ↓ | |
|---|---|---|---|---|---|---|---|---|---|---|---|---|---|---|
| | | | 64-lines | | 32-lines | | 16-lines | | 8-lines | | 4-lines | | | |
| | Abs | RMSE | Abs | RMSE | Abs | RMSE | Abs | RMSE | Abs | RMSE | Abs | RMSE | Abs | RMSE |
| NLSPN* | 0.017 | 3.76 | 0.034 | 4.19 | 0.038 | 4.74 | 0.040 | 5.74 | 0.050 | 7.22 | 0.089 | 10.67 | 2.0 | 2.0 |
| Ours* | 0.014 | 3.67 | 0.023 | 4.08 | 0.026 | 4.62 | 0.030 | 5.51 | 0.045 | 7.03 | 0.080 | 10.01 | 1.0 | 1.0 |

| Methods | Ibims | | NYUv2 | | KITTI | | DIODE | | ETH3D | | Sintel | | Rank ↓ | |
|---|---|---|---|---|---|---|---|---|---|---|---|---|---|---|
| | Abs | RMSE | Abs | RMSE | Abs | RMSE | Abs | RMSE | Abs | RMSE | Abs | RMSE | Abs | RMSE |
| DKN@ | 0.054 | 2.89 | 0.033 | 4.60 | 0.281 | 16.81 | 0.164 | 8.23 | 0.462 | 9.79 | 1.870 | 7.63 | 2.0 | 2.0 |
| Ours@ | 0.025 | 2.06 | 0.023 | 3.89 | 0.066 | 5.70 | 0.134 | 7.47 | 0.134 | 2.40 | 0.253 | 5.32 | 1.0 | 1.0 |

*The table shows the quantitative results on the six test datasets in the three sub-tasks of depth estimation (top), depth completion with different sparsity (center), and depth enhancement (bottom). "#", "\*", and "@" indicate that models are retrained on our augmented dataset with the sampling rates of 0%, (0%, 100%), and 100%, separately.*

TABLE 5. QUANTITATIVE RESULTS OF G2-MONODEPTH AND THE BASELINES ON SYNTHETIC DATA (RETRAINED)

| Sparsity | Methods | Ibims | | NYUv2 | | KITTI | | DIODE | | ETH3D | | Sintel | | Rank | |
|---|---|---|---|---|---|---|---|---|---|---|---|---|---|---|---|
| | | Abs | RMSE | Abs | RMSE | Abs | RMSE | Abs | RMSE | Abs | RMSE | Abs | RMSE | Abs | RMSE |
| 0.1% | NLSPN* | 0.107 | 4.77 | 0.069 | 7.47 | 0.157 | 13.37 | 0.254 | 10.81 | 0.793 | 9.29 | 25.860 | 35.22 | 2.0 | 2.0 |
| | Ours* | 0.059 | 3.27 | 0.049 | 5.85 | 0.134 | 11.61 | 0.230 | 9.98 | 0.435 | 5.73 | 1.689 | 8.31 | 1.0 | 1.0 |
| 1% | NLSPN* | 0.054 | 2.71 | 0.042 | 4.92 | 0.084 | 7.86 | 0.182 | 8.39 | 0.216 | 3.67 | 11.762 | 17.21 | 2.0 | 2.0 |
| | Ours* | 0.031 | 2.24 | 0.028 | 4.14 | 0.064 | 7.00 | 0.163 | 8.06 | 0.133 | 2.53 | 0.610 | 5.69 | 1.0 | 1.0 |
| 10% | NLSPN* | 0.045 | 2.44 | 0.036 | 4.49 | 0.077 | 6.39 | 0.163 | 7.68 | 0.143 | 2.92 | 6.437 | 11.34 | 2.0 | 2.0 |
| | Ours* | 0.026 | 2.08 | 0.024 | 3.87 | 0.060 | 5.77 | 0.142 | 7.44 | 0.076 | 1.87 | 0.338 | 5.18 | 1.0 | 1.0 |

average. Notably, our unified model only uses 0.23 million training data and a basic network with 18M parameters. By comparison, MiDas uses 19 million training data and a network with 105M parameters and it is pretrained on ImageNet with 14000K samples. In the sub-task of depth completion with different sparsity (i.e., 4/8/16/32/64-line), our G2-MonoDepth considerably outperforms all the baselines on all the metrics on average. Indeed, our model achieves the best results in most scenarios. GuideNet and TWISE perform a little better than our model on RMSE and Abs, respectively, in the scenario of 64-line on KITTI. It is because these two models are trained on the training data of KITTI. Notably, the four datasets Ibims, DIODE, ETH3D, and Sintel do not provide real raw depth data for depth completion. The effectiveness of our model will be further verified on the synthetic data of these datasets later. In the sub-task of depth enhancement, our G2-MonoDepth performs much better than all the baselines on the outdoor dataset KITTI and ETH3D, and a little better on the other dataset. The famous JIFF achieves the best RMSE result on NYUv2. However, our model always achieves better visual results than JIFF later.

**Quantitative results on synthetic data.** Most public datasets do not provide real raw depth data for depth completion such as Ibims, DIODE, ETH3D, and Sintel in

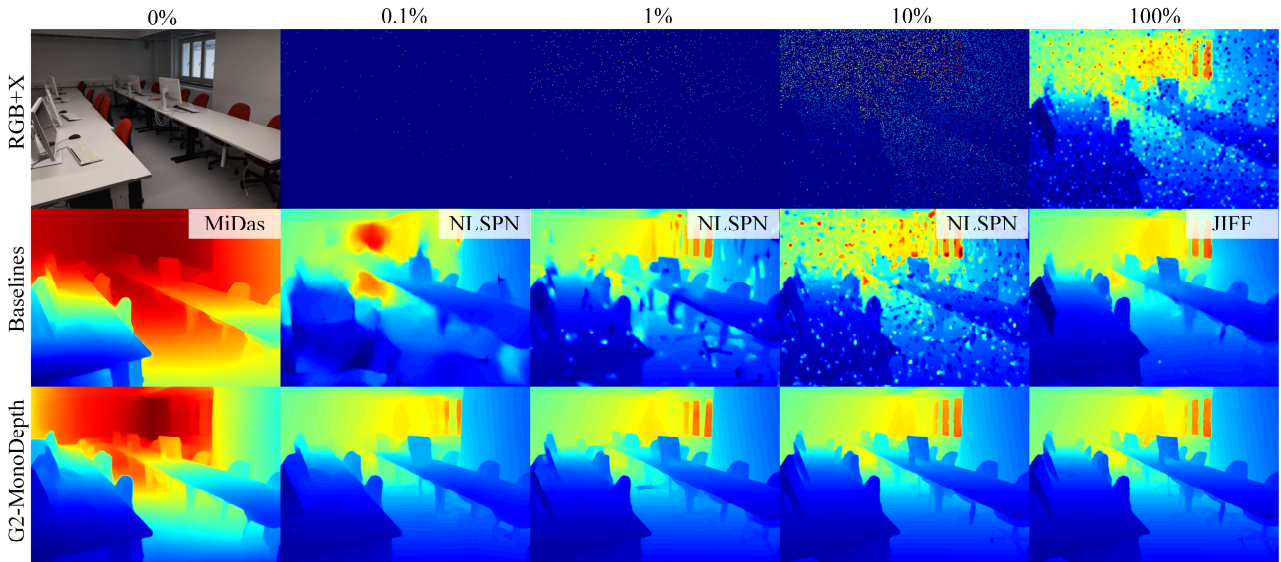

Fig. 6. Visual results on *Ibims*.

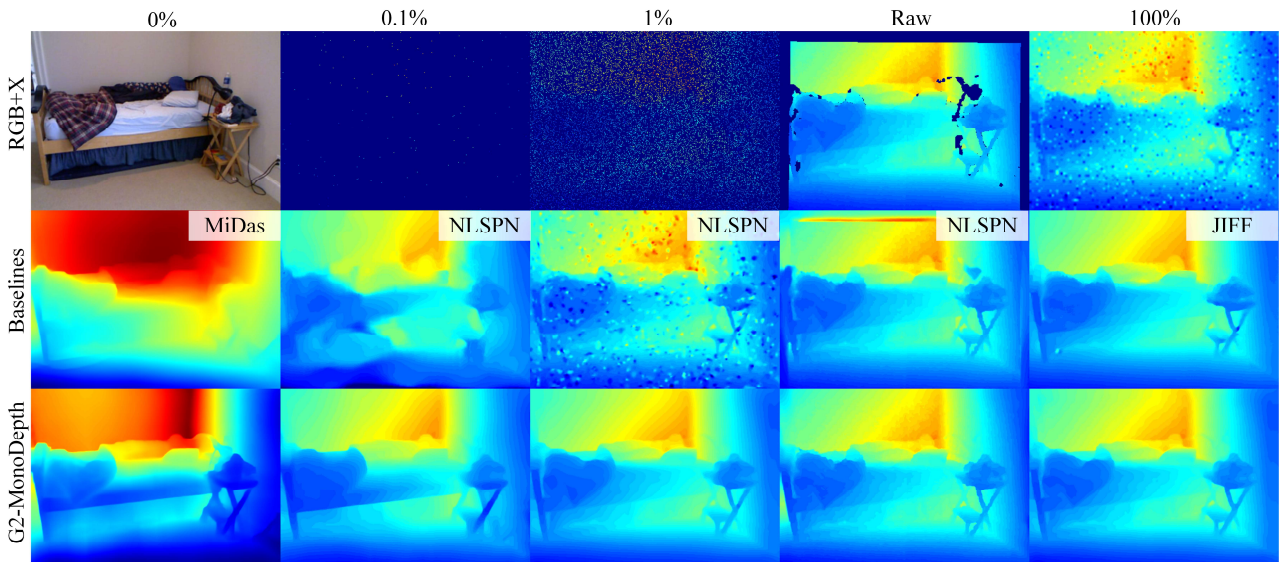

Fig. 7. Visual results on *NYUv2*.

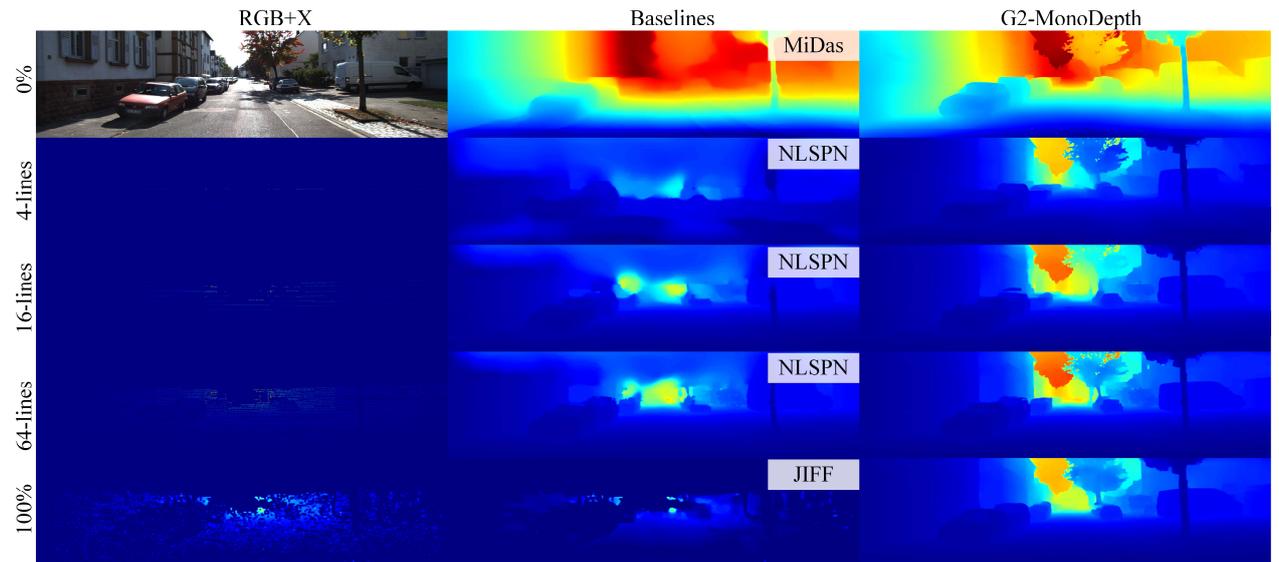

Fig. 8. Visual results on *KITTI*.

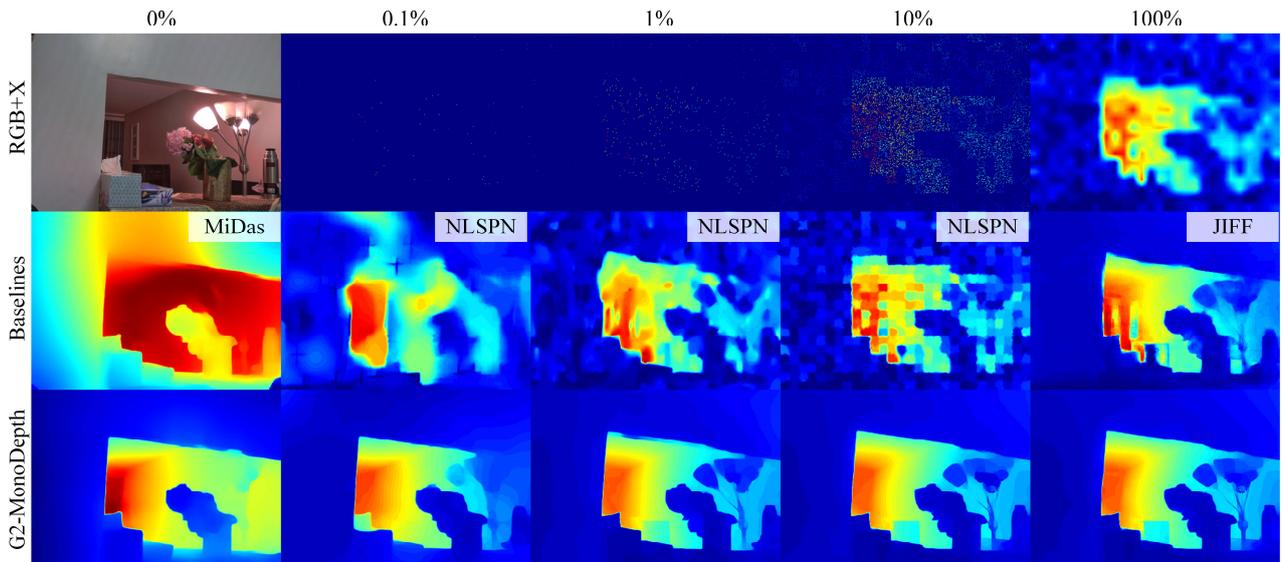

Fig. 9. Visual results on *DIODE*.

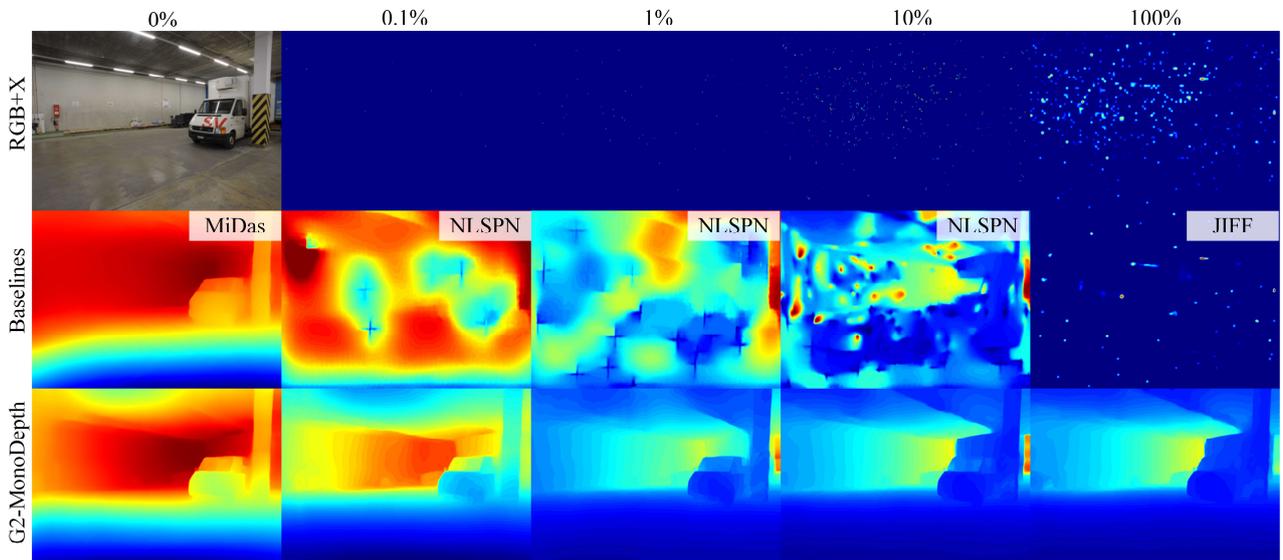

Fig. 10. Visual results on *ETH3D*.

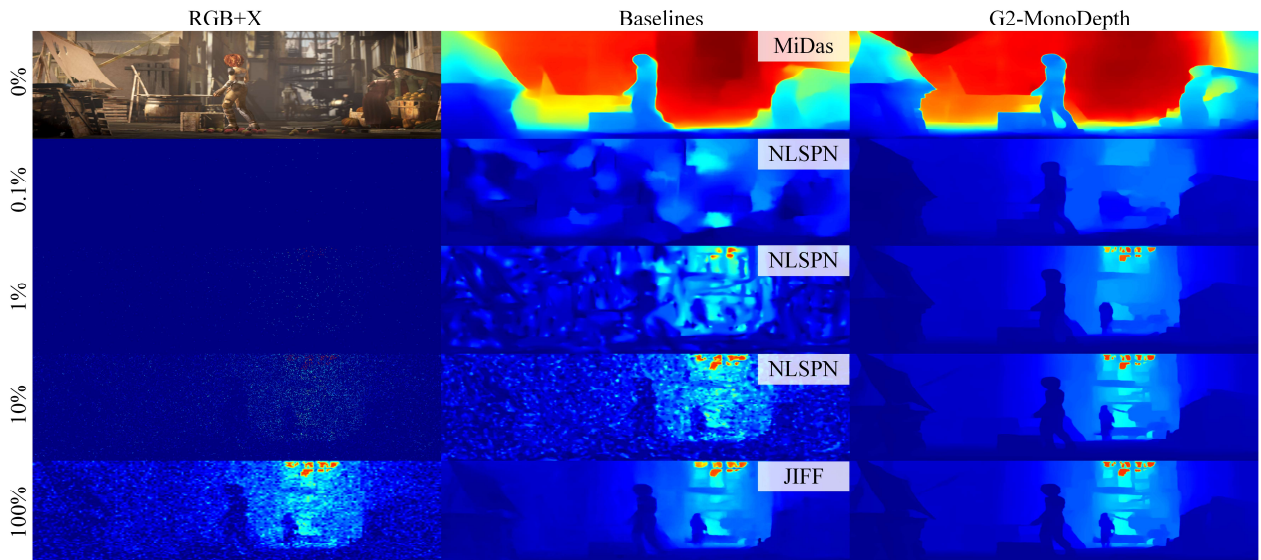

Fig. 11. Visual results on *Sintel*.

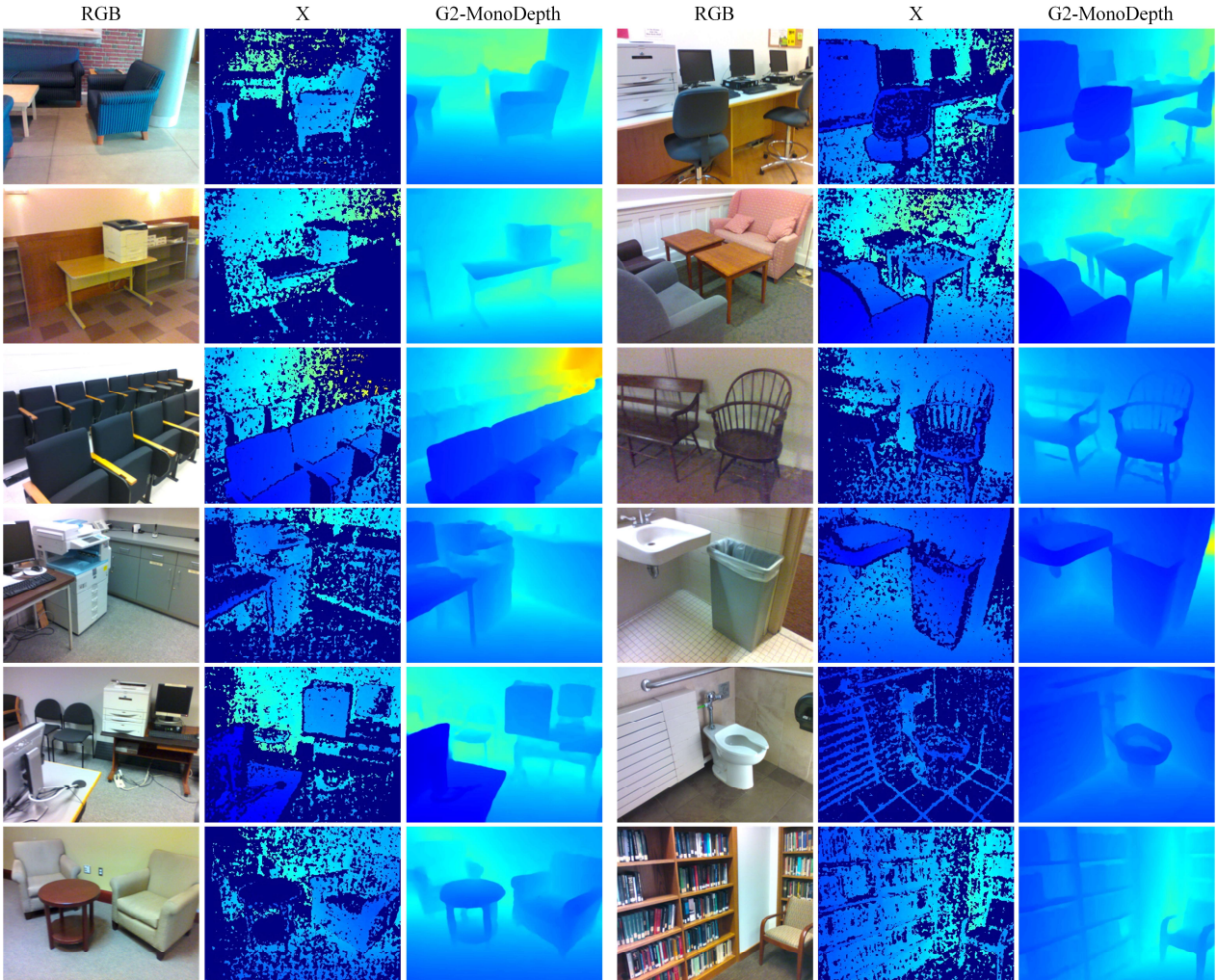

| RGB | X | G2-MonoDepth | RGB | X | G2-MonoDepth |

Fig. 12. Visual results on *Realsense*.

our test. Therefore, we additionally verify the effectiveness of our unified model on synthetic data in the sub-task of depth completion. Table 3 shows the quantitative results of our G2-MonoDepth and the SOTA baselines on the synthetic data with diverse sparsity and errors. Our G2-MonoDepth significantly outperforms the baselines in all the scenarios (i.e., 0.1%, 1%, and 10%) on the six synthetic datasets. The reason lies that the baseline models could not well adapt to the diversity of scene scales/semantics and depth sparsity/errors. Notably, our unified model even outperforms GuideNet and TWISE on the test data of KITTI, though they are trained on the training data of KITTI.

**Retrained as specific models.** Our G2-MonoDepth is used to uniformly handle all the sub-tasks of monocular depth inference. Actually, it can be degraded to three specific models for the three individual sub-tasks in unseen scenes. It is realized by retraining the unified model on our augmented training dataset with the sampling rates of 0%, (0%, 100%), and 100%, separately. For a fair comparison, we similarly retrain the SOTA baseline models of the three sub-tasks on the same training dataset including MiDas, NLSPN, and DKN. Notably, MiDas and NLSPN perform the best in the sub-tasks of depth estimation and depth completion among all the baselines in Table 2. JIFF performs the best in the sub-task of depth enhancement; however, it fails to converge on our training dataset. As a result, we retrain DKN in this sub-task. The results in Tables 4 and 5 show that our specific models still achieve the best results for all the three sub-tasks on both the real-world data and the synthetic data compared with the retrained SOTA baseline models. It further ensures the effectiveness of our unified model to adapt to diverse scene scales/semantics and depth sparsity/errors compared with the baselines.

**Visual results.** Figs. 6-11 show six examples of inferred depth maps from the real-world datasets Ibims, NYUv2, KITTI, DIODE, ETH3D, and Sintel. For simplicity, we only provide visual results of the three SOTA baselines MiDas [16], NLSPN [12], and JIFF [41], which perform the best compared with the other baselines for the three sub-tasks in Table 2. Our G2-MonoDepth always achieves better visual results with clear object boundaries compared with the baseline models in the three sub-tasks. It is worth mentioning that NLSPN fails in some cases due to the diversity of input data in our test, and JIFF does not work well when GT depth maps are sparse or contain holes. Fig. 12 additionally provides the visual results of our model on real sparse depth maps from RealSense.

**Depth sparsity analysis.** Our G2-MonoDepth infers a high-quality depth map from an RGB image and a raw depth map with [0%, 100%] valid pixels in unseen scenes of diverse scales. We observe that the sparsity of raw depth maps significantly affects the scales of inferred depth maps. Fig. 13 shows the experiments on the synthetic data, in which RMSE and SRMSE of inferred depth maps decrease gradually with the increase of the percentage of valid depth pixels from 0% to 100%. More importantly, RMSE drops sharply when the percentage of valid depth pixels increases from 0% to 0.1%, because RMSE measures the absolute relation between inferred depth maps and GT depth maps. In the case of "0%", inferred depth maps often have correct relative relations but wrong scales compared with GT depth maps. These wrong scales of inferred depth maps are corrected when valid depth pixels are additionally provided in the case of "0.1%". By comparison, SRMSE drops slowly from 0% to 0.1%, because it measures the relative relations between inferred depth maps and GT depth maps. It indicates that sparse depth values are crucial to infer the absolute scales of unseen scenes in monocular depth inference. Even a very few sparse depth values are provided, the quality of inferred depth maps will be well improved.

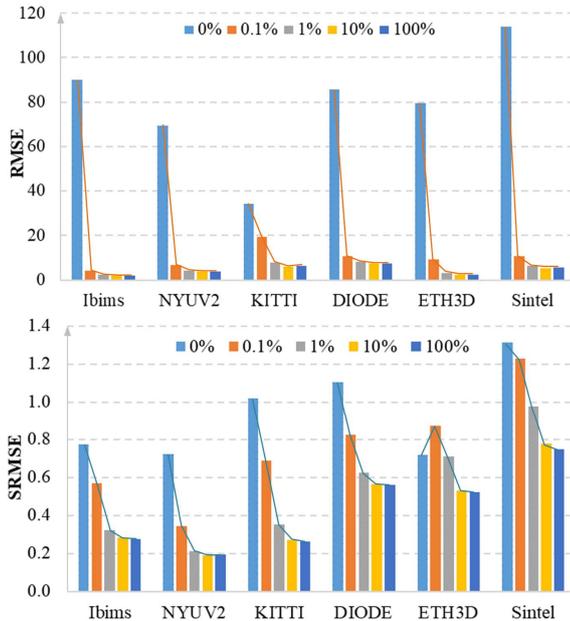

Fig. 13. Effect of depth sparsity on inferred depth maps.

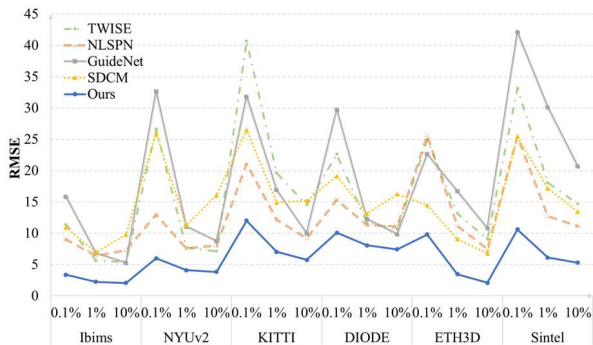

Fig. 14. Stability of G2-MonoDepth across the test datasets with diverse scales/semantics.

**Model stability analysis.** Our G2-MonoDepth aims to infer high-quality depth maps in unseen scenes with diverse scales/semantics. Therefore, we test the stability of G2-MonoDepth across six unseen datasets with different scales/semantics. Fig. 14 shows the RMSE curves across different datasets in the cases of "0.1%, 1%, 10%". Our G2-MonoDepth stably achieves the lowest RMSE across different datasets compared with the baselines. It indicates that G2-MonoDepth is more stable in inferring depth maps in unseen scenes of different scales/semantics.

**Complexity vs. performance.** Our G2-MonoDepth aims to establish a benchmark with a basic backbone and low complexity. Fig. 15 compares our G2-MonoDepth and the baselines in terms of "Complexity vs. Performance". The horizontal coordinate shows the number of network parameters. The vertical coordinate shows the gain of our G2-MonoDepth over the baselines in terms of "Rank" (with RMSE) in Table 2, that is, "Rank" of the baselines minus "Rank" of G2-MonoDepth. Our G2-MonoDepth is located at the bottom-left corner of Fig. 15 compared with the baselines. It indicates that G2-MonoDepth achieves higher performance with fewer network parameters.

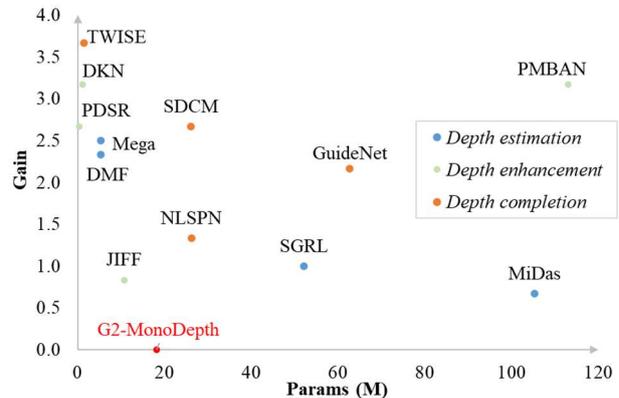

Fig. 15. Complexity vs. performance. The horizontal coordinate shows the number of network parameters. The vertical coordinate shows the gain of our G2-MonoDepth over the baselines in terms of "Rank" (with RMSE) in Table 2.

### 5.3 Ablation Studies

The loss and the network of our G2-MonoDepth are studied in this section. The ablation studies of our model are conducted on the synthetic dataset with more diverse scene scales/semantics and depth sparsity/errors.

**Standardization operations in the loss.** In our G2-MonoDepth loss, we adopt the standardization operation of mean and mean deviation in (4) and (5) (denoted "**G2S**"). G2S can be replaced by other operations, such as mean and standard deviation in z-score standardization (denoted "**ZS**"), and median and median deviation in MiDas [16] (denoted "**MS**"). The ablation study on these three operations is shown in Table 6. The operation MS performs the best in the case of "0%" but the worst in the cases of "0.1%, 1%, 10%, 100%". It even fails in the case of "10%" in NYUv2 and "100%" in KITTI. Oppositely, ZS performs the best in the cases of "0.1%, 1%, 10%, 100%" but the worst in the case of "0%". ZS even introduces a few outliers into inferred depth maps in the case of "0%", such as the example in Fig. 16. These results indicate that

TABLE 6. Ablation study of Three Standardization Operations, ZS, MS, and G2S, in the Loss

| Sparsity | Oper. | Ibims | | NYUv2 | | KITTI | | DIODE | | ETH3D | | Sintel | | Rank ↓ | |
|---|---|---|---|---|---|---|---|---|---|---|---|---|---|---|---|
| | | OE | SRMSE | OE | SRMSE | OE | SRMSE | OE | SRMSE | OE | SRMSE | OE | SRMSE | OE | SRMSE |
| 0% | ZS | **0.207** | 0.92 | 0.195 | 0.83 | 0.171 | <u>0.99</u> | <u>0.245</u> | 1.23 | 0.192 | 0.86 | **0.333** | 1.36 | <u>2.0</u> | 2.8 |
| | MS | 0.212 | **0.78** | 0.190 | **0.70** | 0.158 | **0.87** | 0.255 | <u>1.14</u> | <u>0.186</u> | **0.76** | 0.342 | **1.30** | 2.0 | **1.3** |
| | G2S | <u>0.211</u> | 0.83 | <u>0.194</u> | <u>0.75</u> | <u>0.167</u> | 1.03 | 0.245 | 1.11 | **0.176** | **0.73** | 0.348 | <u>1.33</u> | 2.0 | <u>1.8</u> |
| | | Abs | RMSE | Abs | RMSE | Abs | RMSE | Abs | RMSE | Abs | RMSE | Abs | RMSE | Abs | RMSE |
| 0.1% | ZS | **0.035** | 2.64 | **0.045** | 7.48 | 0.526 | <u>29.72</u> | 0.194 | 9.81 | <u>0.458</u> | <u>6.42</u> | **1.295** | **8.66** | **1.2** | **1.2** |
| | MS | 0.052 | 3.14 | 0.059 | 8.67 | <u>0.527</u> | **29.36** | 0.223 | <u>10.22</u> | 0.622 | 7.49 | <u>1.790</u> | 10.30 | 2.7 | 2.5 |
| | G2S | <u>0.038</u> | <u>2.81</u> | <u>0.050</u> | <u>8.30</u> | 0.595 | 31.87 | <u>0.222</u> | 10.49 | **0.443** | 6.42 | 2.312 | <u>9.99</u> | <u>2.2</u> | <u>2.3</u> |
| 1% | ZS | **0.017** | **1.80** | **0.024** | **4.76** | 0.068 | **9.84** | 0.138 | 7.68 | 0.065 | **1.60** | **0.242** | **5.14** | **1.0** | **1.0** |
| | MS | 0.024 | 2.05 | 0.041 | 9.82 | 0.121 | 11.30 | <u>0.146</u> | <u>7.94</u> | **0.096** | 2.01 | <u>0.747</u> | 5.89 | <u>2.5</u> | 2.8 |
| | G2S | <u>0.022</u> | <u>1.82</u> | <u>0.027</u> | <u>5.30</u> | <u>0.103</u> | <u>10.61</u> | 0.149 | 7.95 | <u>0.096</u> | <u>1.78</u> | 0.809 | <u>5.58</u> | 2.5 | <u>2.2</u> |
| 10% | ZS | **0.010** | **1.16** | 0.018 | 4.02 | <u>0.032</u> | 5.67 | 0.098 | 5.65 | 0.019 | 0.61 | 0.143 | **3.16** | **1.2** | **1.0** |
| | MS | <u>0.014</u> | 1.26 | 0.063 | 15.75 | 0.037 | 5.81 | <u>0.105</u> | 5.83 | **0.031** | 0.84 | <u>0.194</u> | 3.26 | <u>2.3</u> | 2.8 |
| | G2S | 0.017 | <u>1.21</u> | <u>0.020</u> | <u>4.28</u> | **0.031** | <u>5.72</u> | 0.106 | <u>5.76</u> | 0.039 | <u>0.72</u> | 0.259 | <u>3.29</u> | 2.5 | <u>2.2</u> |
| 100% | ZS | 0.025 | 2.26 | 0.021 | <u>3.91</u> | <u>0.052</u> | <u>8.95</u> | 0.126 | 7.86 | <u>0.061</u> | <u>2.10</u> | **0.151** | <u>5.32</u> | **1.7** | **1.7** |
| | MS | <u>0.025</u> | 2.29 | <u>0.021</u> | **3.88** | 0.288 | 18.98 | 0.135 | 8.04 | 0.192 | 5.60 | <u>0.269</u> | **5.26** | <u>2.2</u> | 2.3 |
| | G2S | 0.029 | <u>2.28</u> | 0.022 | 3.98 | **0.041** | **7.49** | <u>0.132</u> | <u>7.87</u> | **0.055** | **1.74** | 0.323 | 5.40 | <u>2.2</u> | <u>2.0</u> |

TABLE 7. Ablation Study of Gradient Operators, Diff. and Sobel, in the Loss

| Sparsity | Oper. | Ibims | | NYUv2 | | KITTI | | DIODE | | ETH3D | | Sintel | | Rank ↓ | |
|---|---|---|---|---|---|---|---|---|---|---|---|---|---|---|---|
| | | OE | SRMSE | OE | SRMSE | OE | SRMSE | OE | SRMSE | OE | SRMSE | OE | SRMSE | OE | SRMSE |
| 0% | Diff. | **0.200** | **0.73** | 0.188 | 0.72 | 0.159 | **1.00** | 0.238 | 1.10 | **0.176** | 0.72 | 0.338 | 1.31 | **1.2** | **1.2** |
| | Sobel | 0.208 | 0.77 | 0.193 | **0.72** | **0.158** | 1.02 | 0.244 | **1.10** | 0.179 | 0.72 | 0.343 | 1.31 | 1.8 | 1.8 |
| | | Abs | RMSE | Abs | RMSE | Abs | RMSE | Abs | RMSE | Abs | RMSE | Abs | RMSE | Abs | RMSE |
| 0.1% | Diff. | **0.037** | 2.77 | 0.047 | 7.56 | **0.511** | 29.66 | 0.201 | 10.13 | **0.508** | 6.78 | 1.912 | 9.20 | **1.2** | 1.7 |
| | Sobel | 0.039 | **2.63** | **0.046** | 7.15 | 0.565 | 31.32 | 0.212 | 9.90 | 0.558 | 7.68 | 1.937 | 8.98 | 1.8 | **1.3** |
| 1% | Diff. | **0.017** | 1.75 | 0.025 | 4.97 | **0.101** | 10.96 | 0.138 | 7.76 | 0.080 | 1.74 | 0.425 | 5.21 | **1.3** | 2.0 |
| | Sobel | 0.019 | **1.63** | 0.026 | **4.84** | 0.113 | 9.75 | 0.144 | 7.41 | 0.078 | **1.66** | 0.325 | 4.89 | 1.7 | **1.0** |
| 10% | Diff. | 0.009 | 1.13 | 0.019 | 4.18 | **0.031** | 5.46 | 0.098 | 5.70 | 0.022 | 0.63 | 0.134 | 3.26 | **1.3** | 1.8 |
| | Sobel | 0.011 | **1.07** | 0.018 | **3.99** | 0.033 | **5.18** | 0.102 | **5.43** | 0.028 | 0.64 | 0.127 | **2.91** | 1.7 | **1.2** |
| 100% | Diff. | 0.024 | 2.19 | 0.021 | 3.89 | **0.039** | 5.68 | 0.127 | 7.70 | 0.053 | 1.47 | 0.194 | 5.30 | 1.5 | 2.0 |
| | Sobel | 0.024 | **2.08** | 0.021 | **3.74** | 0.041 | **5.19** | 0.128 | 7.40 | 0.053 | 1.41 | 0.181 | **5.09** | 1.5 | **1.0** |

TABLE 8. Ablation Study of the Basic Blocks, BN and ReZero, in the Network

| Sparsity | Blocks | Ibims | | NYUv2 | | KITTI | | DIODE | | ETH3D | | Sintel | | Rank ↓ | |
|---|---|---|---|---|---|---|---|---|---|---|---|---|---|---|---|
| | | OE | SRMSE | OE | SRMSE | OE | SRMSE | OE | SRMSE | OE | SRMSE | OE | SRMSE | OE | SRMSE |
| 0% | BN | 0.152 | 0.58 | 0.165 | 0.65 | **0.150** | 0.95 | 0.214 | 1.02 | 0.161 | **0.65** | **0.309** | **1.21** | 1.7 | 1.7 |
| | ReZero | **0.145** | **0.56** | **0.157** | 0.61 | 0.156 | **0.95** | 0.208 | **1.00** | 0.150 | 0.66 | 0.310 | 1.26 | **1.3** | **1.3** |
| | | Abs | RMSE | Abs | RMSE | Abs | RMSE | Abs | RMSE | Abs | RMSE | Abs | RMSE | Abs | RMSE |
| 0.1% | BN | 0.095 | 3.71 | 0.060 | 6.49 | 0.255 | 16.46 | 0.330 | 11.59 | 2.102 | 25.25 | 3.910 | 12.10 | 2.0 | 2.0 |
| | ReZero | **0.078** | **3.40** | **0.051** | **6.02** | **0.151** | **12.03** | **0.266** | **10.10** | **0.912** | **9.84** | **3.193** | **10.63** | **1.0** | **1.0** |
| 1% | BN | 0.054 | 2.38 | 0.032 | 4.18 | 0.083 | 7.69 | 0.212 | 8.57 | 0.378 | 4.47 | 1.882 | 7.03 | 2.0 | 2.0 |
| | ReZero | **0.035** | **2.28** | **0.029** | **4.14** | **0.066** | **7.04** | **0.173** | **8.10** | **0.262** | **3.47** | **1.060** | **6.14** | **1.0** | **1.0** |
| 10% | BN | 0.041 | 2.14 | 0.027 | 3.92 | 0.062 | 6.00 | 0.168 | 7.56 | 0.160 | 2.48 | 0.791 | 5.61 | 2.0 | 2.0 |
| | ReZero | **0.028** | **2.09** | **0.025** | **3.86** | **0.055** | **5.79** | **0.146** | **7.46** | **0.104** | **2.13** | **0.506** | **5.32** | **1.0** | **1.0** |
| 100% | BN | 0.033 | 2.12 | 0.025 | **3.85** | 0.095 | 6.40 | 0.150 | 7.49 | **0.106** | 2.25 | 0.622 | 5.53 | 1.8 | 1.8 |
| | ReZero | **0.027** | **2.11** | **0.024** | 3.87 | **0.067** | **5.78** | **0.139** | **7.40** | 0.109 | **2.16** | **0.459** | **5.40** | **1.2** | **1.2** |

the two operations, MS and ZS, are not stable in all the cases. By comparison, G2S is always stable in all the cases, though it only performs second-best in Table 6. Fig. 16 shows the visual results of standardization operations.

**Gradient operators in the loss.** In our G2-MonoDepth loss, we adopt the Sobel operator in the gradient term (9) instead of the traditional differential operator [16] [17] [19]. The ablation study on these two gradient operators is shown in Table 7. Overall, the quantitative results of inferred depth maps with these two operators are comparable. The Sobel operator performs slightly better than the differential operator in the cases of "0.1%, 1%, 10%, 100%" (with RMSE). The differential operator performs slightly better in the cases of "0%" and "0.1%, 1%, 10%, 100%" (with Abs). However, the Sobel operator always achieves better visual results with clear boundaries compared with the differential operator in Fig. 16.

**Basic blocks in the network.** In ReZero U-Net, we adopt the ReZero block instead of the widely-used BN block in Fig. 3. The ablation study on these two blocks is shown in Table 8. ReZero block always performs better than BN block. One reason lies in that the BN block may hinder the scale propagation for G2-MonoDepth, which has been analyzed in Section 4. Fig. 16 further shows that ReZero block always achieves better visual results with clear object boundaries compared with BN block.

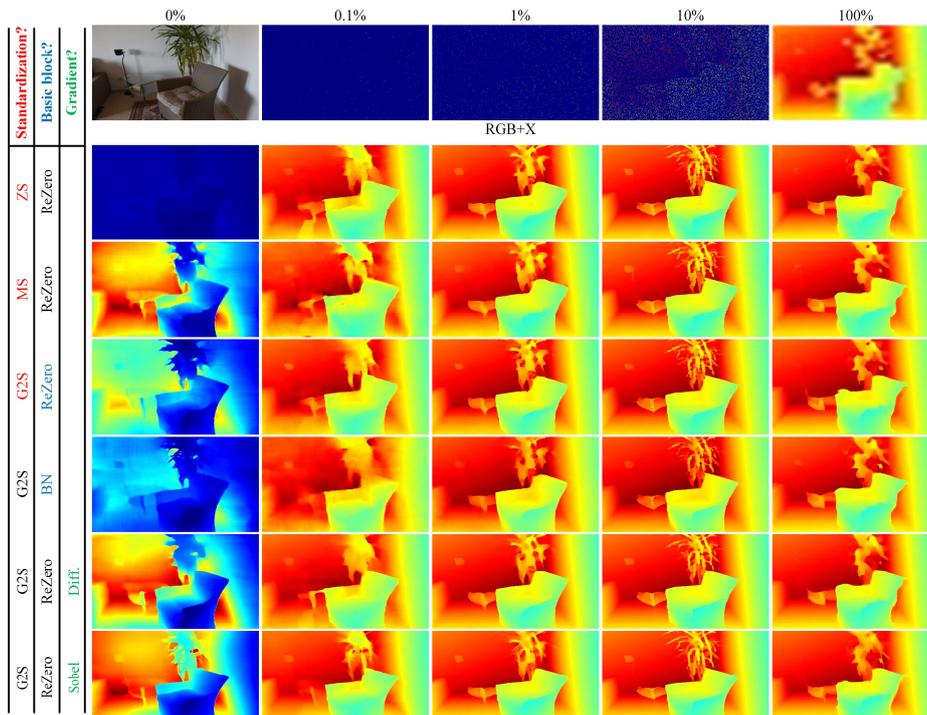

Fig. 16. Visual results of inferred depth maps in the three ablation studies including standardization operations (in red), basic blocks (in blue), and gradient operators (in green).

## 6 CONCLUSIONS

This paper first investigated a unified task of monocular depth inference for various robots that may be equipped with a single camera plus an optional depth sensor of any type and located in unseen scenes of different scales. A basic benchmark, namely G2-MonoDepth, was developed for this unified task. It naturally accommodates the individual sub-tasks of depth estimation, depth completion with different sparsity, and depth enhancement in unseen scenes in the literature. The experiments on both real-world data and synthetic data verified that G2-MonoDepth always achieves competitive results in the three sub-tasks compared with relevant baselines.

Though our benchmark model has achieved desired results in the unified task, three problems remain not fully solved which may help us further improve the accuracy of the model in future work. First, our basic architecture based on ReZero may not well converge for large-scale networks with a large number of parameters [66]. How to design a network architecture to simultaneously propagate scene scales from input to output and ensure easy convergence in the training? Second, our model is trained on synthetic data rather than real-world data because most public RGB-D datasets do not provide sufficient raw depth maps with diverse sparsity/errors for supervised learning. How to address this data limitation issue in an un-supervised or semi-supervised way [42]? Third, existing models with the input of a single RGB image (e.g., monocular depth estimation) usually benefit from pretrained backbones of image classification task. How to develop a pretraining method for our unified model with the input of RGB-D data?

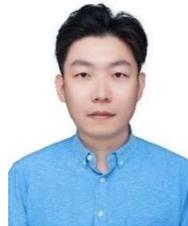

**Haotian Wang** received the B.S. degree in electronic and information engineering in North China Electric Power University in 2017. He is currently pursuing the Ph.D. degree with the Institute of Artificial Intelligence and Robotics, Xi'an Jiaotong University. His research interests include 3D vision and multi-modal vision.

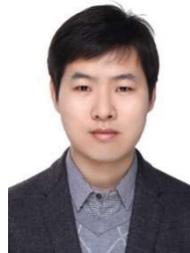

**Meng Yang** (Member, IEEE) received the B.S. degree in information engineering and the Ph.D. degree in control science and engineering, Xi'an Jiaotong University, China, in 2008 and 2014, respectively. He was a visiting scholar with the University of California at San Diego, USA, from 2011 to 2012. He is currently an Associate Professor with the Institute of Artificial Intelligence and Robotics, Xi'an Jiaotong University. His research interests include machine vision and autonomous vehicle.

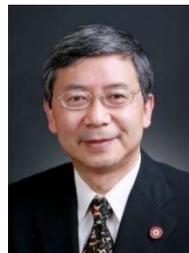

**Nanning Zheng** (Fellow, IEEE) graduated from the Department of electrical engineering, received the M.S. degree in information and control engineering from Xi'an Jiaotong University in 1975 and 1981, respectively, and the Ph.D. degree in electrical engineering from Keio University in 1985. He is currently a Professor with the Institute of Artificial Intelligence and Robotics, Xi'an Jiaotong University. His research interests include computer vision, pattern recognition, autonomous vehicle, and brain-inspired computing. He became a member of the Chinese Academy of Engineering in 1999. He is a Council Member of the International Association for Pattern Recognition. He is the President of the Chinese Association of Automation and the Vice-President of the Chinese Society for Cognitive Science.